\newcommand{\sxh}[1]{\textcolor{black}{#1}}
\newcommand{\fzl}[1]{\textcolor{black}{#1}}
\documentclass[sigconf]{acmart}
\usepackage{multirow} 
\usepackage{array}  
\newcommand{\etal}{\textit{et al.}}
 
\AtBeginDocument{%
  \providecommand\BibTeX{{%
    \normalfont B\kern-0.5em{\scshape i\kern-0.25em b}\kern-0.8em\TeX}}}

\setcopyright{acmlicensed}
\copyrightyear{2020}
\acmYear{2020}
\acmDOI{10.1145/3394171.3413955}

\acmConference[MM '20] {28th ACM International Conference on Multimedia}{October 12--16, 2020}{Seattle, WA, USA.}
\acmBooktitle{28th ACM International Conference on Multimedia (MM '20), October 12--16, 2020, Seattle, WA, USA.}
\acmPrice{15.00}
\acmISBN{978-1-4503-7988-5/20/10}

\begin{document}
\fancyhead{}

\title[]{Unsupervised Learning Facial Parameter Regressor for Action Unit Intensity Estimation via Differentiable Renderer
}

\author{Xinhui Song$^{1}$, Tianyang Shi$^{1}$, Zunlei Feng$^{2}$, Mingli Song$^{2}$}
\author{Jackie Lin$^{1}$, Chuanjie Lin$^{1}$, Changjie Fan$^{1}$, Yi Yuan$^{1*}$}
\affiliation{\institution{$^1$NetEase Fuxi AI Lab ($^*$ Corresponding author)}}
\affiliation{\institution{$^2$Zhejiang University}}
\email{{songxinhui, shitianyang, linjie3, linchuanjie, fanchangjie, yuanyi}@corp.netease.com, {zunleifeng, brooksong}@zju.edu.cn}


\begin{abstract}
Facial action unit (AU) intensity is an index to describe all visually discernible facial movements. Most existing methods learn intensity estimator with limited AU data, while they lack of generalization ability out of the dataset. In this paper, we present a framework to predict the facial parameters (including identity parameters and AU parameters) based on a bone-driven face model (BDFM) under different views. The proposed framework consists of a feature extractor, a generator, and a facial parameter regressor. The regressor can fit the physical meaning parameters of the BDFM from a single face image with the help of the generator, which maps the facial parameters to the game-face images as a differentiable renderer. Besides, identity loss, loopback loss, and adversarial loss can improve the regressive results. Quantitative evaluations are performed on two public databases BP4D and DISFA, which demonstrates that the proposed method can achieve comparable or better performance than the state-of-the-art methods. What’s more, the qualitative results also demonstrate the validity of our method in the wild.
\end{abstract}

\ccsdesc[500]{Applied computing~Computer games}
\ccsdesc[300]{Computing methodologies~Computer vision}

\keywords{AU Intensity Estimation; Differentiable Renderer}

\maketitle
\begin{figure} 
	\includegraphics[width=0.46\textwidth]{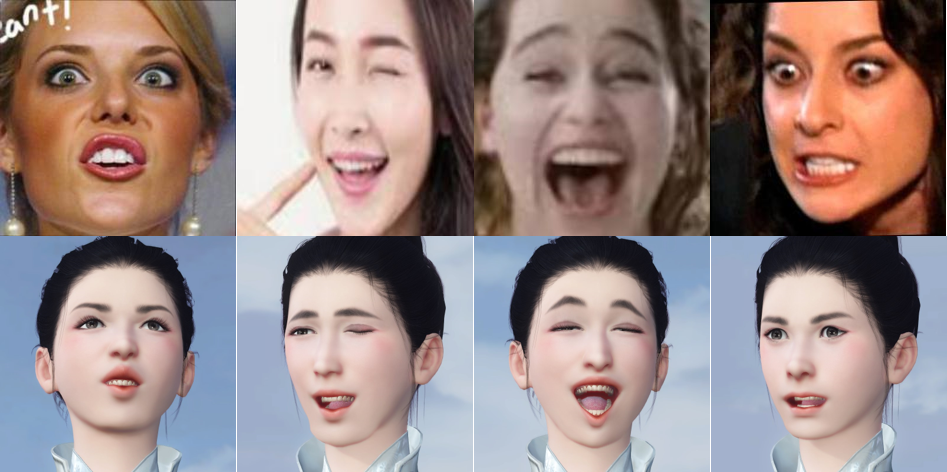}
	\caption{
		Visual results of our method. First row: face images with different expressions and different views; Second row: 3D characters generated with the \sxh{predicted} facial parameters, which are learned from the face images. \sxh{Original photos in the first row are courtesy of Carrie Prejean, piotr marcinski, Emilia Clarke and Sin, respectively}. 
	}
	\Description{Results of our approach.}
	\label{overview}
\end{figure}

\section{Introduction}
Facial expression analysis is a research hotspot in computer vision and computer graphics.
It can be useful in many applications, such as human-robot interaction, social interaction analysis, and medical treatments.
Facial expression analysis includes many research tasks: AU intensity estimation, 3D face reconstruction, facial expression recognition, facial action unit detection and so on.
Among them, the AU intensity estimation \cite{friesen1978facial} is the most challenging task, which aims to estimate anatomically meaningful parameters (i.e., muscle movements). 
Ekman and Friesen~\cite{friesen1978facial} develop a Facial Action Coding System (FACS) \sxh{to characterize human expressions}, which refers to a set of facial muscle movements that correspond to a displayed emotion.
Nearly any anatomically possible facial expression can be coded by a combination of AUs. 
The difficulty of the AU intensity estimation steams from both the fine-grained analysis and meaningful parameters separated from different individuals.
That is to say, the different individuals can dramatically increase variances of face images even when they have the same expression.
Besides, pose also brings a challenge. 
Recently, some AU intensity estimation methods \cite{walecki2016copula,kaltwang2015latent,rudovic2014context, zhao2016facial,zhang2018weakly,zhang2018bilateral,zhang2019context,zhang2019joint} require a large set of labeled training corpus, whether it is a fully supervised method or weakly supervised method.
However, AU annotation requires strong domain expertise, resulting in very high time-and-labor costs to construct 
\fzl{an extensive database}.

3D face reconstruction is used to estimate a personalized human face model from a single photo.
Recently, it is expanded to describe facial shapes and expressions.
Many representative methods model the human face by estimating the \sxh{coefficients} 
of the 3D Morphable Models (3DMM) \cite{blanz1999morphable} or the 3D Based Face Model \cite{bfm09}.
The 3D models are obtained from scan databases of limited size so that the faces are constrained to lie in a low-dimensional subspace.
\sxh{Whereas, some algorithms do not generalize well beyond the restricted low-dimensional subspace of the 3DMM.}

\fzl{The bone-driven face model we adopted and used in the existing works~\cite{F2p, genet2020} is a character face rendering model of the game \emph{Justice}\footnote{https://n.163.com/}. Meanwhile, similar 
face models are popularly used in many role-playing games (RPGs) (e.g., ``Grand Theft Auto V", ``Dark Souls III" and ``A Dream of Jianghu") }.
The facial parameters of the BDFM are the facial component offsets to the base face and have explicit physical meanings.
The ``base face" is created by rendering a frontal emotionless face image, with the identity parameters set to 0.5.
The identity parameters create a neutral face for 
\fzl{an} identity-specific face shape by controlling the attributes of facial components, including position, orientation, and scale.
The offsets relative to the neutral face are baked to AU parameters, which can drive the facial muscle movement. 
The AU parameters can express almost all the AUs in FACS.
They train a generator to make the bone-driven face model differentiable, and the facial reconstruction is formulated as an optimization problem.
The iterative optimization process updates the facial parameters by minimizing the distance of the generated image and the input photo.  
However, the method has some limitations, including robustness and processing speed.  
\fzl{Besides, the bone-driven face model has other additional challenges that some identity parameters corresponding images are meaningless and identity parameters tangle with AU parameters in some cases.}

\begin{figure*} 
	\includegraphics[width=\linewidth]{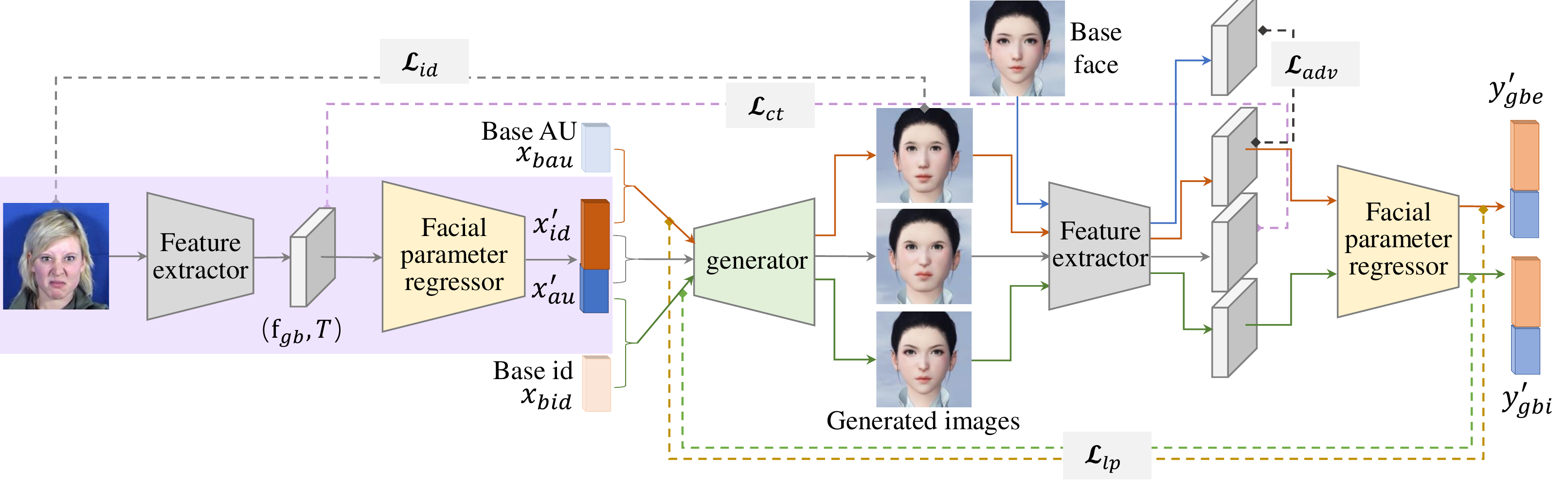}
	\caption{The learning framework for regressing meaningful facial parameters.
		The framework consists of four components: a feature extractor, a facial parameter regressor, a generator and a facial identity recognition network (not shown in the figure).  
		The feature extractor is used to extract the facial features $(f_{gb},T)$. 
		The facial parameter regressor is utilized to regress the facial parameters from the facial features.
		The generator is a differentiable renderer for mapping the facial parameters to the rendered image. 
		\sxh{In the training phase, only the network parameters of the facial parameter regressor are learnable, and they are updated by computing multiple losses with the generated images.}
		The different colored lines mean the different flows. Only the part of the purple background is used in the predicting phase. Original image courtesy of MMI dataset.
	}
	\label{framework_overview}
\end{figure*}

To solve the above problems, we present a trainable method to learn a facial parameter regressor $\mathcal{G}$  with the help of a feature extractor $F_{seg}$, a generator and a face recognition network $F_{reg}$.
We can train an end-to-end framework
by using the rendered images to supervise the facial parameters fitting under the self-supervised paradigm. 
The inference speed is improved by two orders of magnitude than the optimization method.
The overview of the proposed framework is shown in Fig. \ref{framework_overview}.
The feature extractor $F_{seg}$ extracts the shape-sensitive features.
The facial parameter regressor is exploited to fit the facial parameters from the facial representation instead of the raw pixel image.
The generator imitates the rendering process of game engines and makes the rendering process differentiable as introduced by previous methods \cite{F2p, genet2020}.


In the training phase, the network parameters of all networks are fixed except for the facial parameter regressor.
The total loss function consists of five parts: facial content loss, identity loss, parameter loss, adversarial loss, and loopback loss.
\fzl{We utilize the facial content loss to measure the pixel-wise distance between the input face image and image generated with the predicted face parameters.}
The identity loss of the face recognition network $F_{reg}$ can constrain the facial parameter regressor to match the identity features. 
The parameter loss and the adversarial loss are adopted to 
\fzl{constrain the sparsity and disentanglement of the facial parameters, respectively.}
The loopback loss ensures the facial parameters regressor correctly interpret its output.

Our contributions are summarized as the following:
\sxh{
1) The first unsupervised method that estimates AU intensity from a single natural image without the AU annotations.
2) A pre-trained deep model that maps the facial parameters to a 3D character is incorporated into the framework to realize a differentiable framework. 
3) Meanwhile, a disentangling mechanism is proposed to learn the mapping between the natural image and the facial parameters.
4) The proposed method achieves real-time prediction and SOTA performance.
}

\section{Related Works}
{\bf Learning to Regress 3D Face Models.}
Most methods \cite{cao2014displaced,garrido2016reconstruction,tuan2017regressing,tewari2018self,jiang2019disentangled,chaudhuri2019joint,yi2019mmface} learn the coefficients of the 3D morphable model (3DMM) \cite{blanz1999morphable} to fit the input face image.
\fzl{Some} previous approaches \cite{cao2014displaced,garrido2016reconstruction} fit the 3DMM from a single face image by solving an optimization problem. 
However, these methods require expensive computation; moreover, they are sensitive to initialization.  
Deep learning methods \cite{tuan2017regressing,tewari2018self,jiang2019disentangled,chaudhuri2019joint,yi2019mmface} show the ability in regressing the 3DMM coefficients from a face image. 
However, these methods require an extensive collection of annotated training data, which is limited by 3D scan databases.
One solution is to generate synthetic training data by sampling randomly from the 3DMM and rendering the corresponding faces \cite{richardson20163d}.
Whereas, a deep network trained on purely synthetic data may present poorly due to the data is in limited condition. 
\fzl{Other methods obtain the training data with an iterative optimization method, where the final optimized results are set as ground truth. However, the performance of those methods is always limited by the expressiveness of the iterative optimization method and the accuracy of the fitting process.}
For example, Tran \etal~\cite{tuan2017regressing} propose a deep ResNet to regress the same 3DMM coefficients for different photos of the same subject. 
Due to the lack of training data, some unsupervised learning method are developed.
For instance, Sanyal \etal~\cite{sanyal2019learning} use RingNet to learn the 3D face shape. The RingNet leverages multiple images of a subject to train the model, which encourages the face shape to be similar when the identity is the same and different for different people. 
MOFA \cite{tewari2017mofa} contains a convolutional encoder network and an expert-designed generative model. The generative model is used to render the face with the 3DMM coefficients.
Genova \etal~\cite{genova2018unsupervised} utilize the differentiable rasterizer to form neutral-expression face images.
Inspired by the unsupervised learning methods, our approach can regress the meaningful facial parameters based on the differentiable rendering convolutional network without annotated data.

{\bf Supervised AU Intensity Estimation.}
Up to now, most of the works on automated AU analysis focus on AU detection. However, AU detection is a binary classification problem, so the methods can not estimate the AU intensity.
AU intensity estimation is a more challenging task than AU detection and is a relatively new problem in the field.
Recently, few works try to address it.
Many \cite{sandbach2013markov,ming2015facial,jeni2013continuous,kaltwang2015doubly,kaltwang2012continuous} of these works independently estimate the AU intensity without considering the relationships among AUs.
\fzl{Considering the relationships, many researchers exploit the probabilistic graphical models \cite{walecki2016copula,kaltwang2015latent,rudovic2014context}.}
Walecki \etal~\cite{walecki2016copula} use Copula Ordinal Regression (COR) to estimate the intensities of multiple AUs.
Kaltwang \etal~\cite{kaltwang2015latent} exploit a latent tree model by conducting a graph-edits for representing the joint distribution of targets and features. 
Besides, temporal continuous is used in hidden Markov model \cite{mavadati2014temporal} and continuous conditional neural fields (CCNF) \cite{baltruvsaitis2014continuous}.
Recently, several deep learning methods are proposed for AU intensity estimation, including CNN \cite{gudi2015deep}, CNN-IT \cite{walecki2017deep} and 2DC \cite{linh2017deepcoder}.
These supervised learning methods require a large amount of annotated images for training.
However, annotating AU intensity is very difficult, which requires strong domain expertise. So it is expensive and laborious to construct 
\fzl{an extensive database}.
With existing limited AU annotation data, supervise methods have two shortcomings: overfitting on the training set and limited AU expressions.
\fzl{However, our approach can estimate the AU intensity without AU annotated data and extend the number of estimable AU. }

{\bf Weakly Supervised AU Intensity Estimation.}
A few weakly supervised methods \cite{zhao2016facial,zhang2018weakly,zhang2018bilateral,zhang2019context,zhang2019joint} use partially labeled images to train AU intensity estimation models.
Zhao \etal~\cite{zhao2016facial} exploit the ordinal information to train a regressor with labeled and unlabeled frames among an expression sequence.
Zhang \etal~\cite{zhang2018weakly} propose a knowledge-based deep convolutional neural network for AU intensity estimation with peak and valley frames in training sequences.
BORMIR \cite{zhang2018bilateral} formulates the AU intensity estimation as a multi-instance regression problem with weakly labeled sequences.
Zhang \etal~\cite{zhang2019context} present a patch-based deep model 
\fzl{consisting of} a feature fusion module and a label fusion module.
Zhang \etal~\cite{zhang2019joint} joint learn representation and intensity estimator to achieve an optimal solution with multiple types of human knowledge. 
However, these methods still require high-quality annotated training corpus. And the human-defined knowledge restricts the space of anatomically plausible AUs.
Differently, our method requires neither a large scale of data nor restricts the space of AUs.
It \fzl{can still} accurately regress multiple AUs with the help of the differentiable renderer.

\subsection{Method}

We expand an encoder-decoder architecture that permits learning of the meaningful facial parameters (Fig. \ref{framework_overview}).
\fzl{The} training framework consists of four sub-networks: a feature extractor $F_{seg}$, a facial parameters regressive network $\mathcal{R}$, a facial generative network $\mathcal{G}$, and a face recognition network $F_{reg}$. 
\fzl{The deep features extracted by the feature extractor are the input of the facial parameters regressive network, which regresses the facial parameters. }
\fzl{The facial generative network is a differentiable renderer, which renders expressive face images with the facial parameters under varying pose.}
The rendered face image is the input of the face recognition network and the feature extractor, 
\fzl{which supervises the training of the facial parameters regressor.}
\fzl{Only the facial parameter regressor's parameters are updated in the training phase, while the pre-trained feature extractor is fixed.}

\subsection{Generator}
\sxh{In the role-playing games, the parameterized physical model (the bone-driven face 3D model) is usually adopted to render a 3D character image. Each parameter of the parameterized model can control the facial component offset. 
However, the parameterized physical model is not differentiable. To make the whole framework differentiable, we train a generator to imitate the render process with the paired facial parameters and the rendered images.}
The generator $\mathcal{G}: y_g  \longrightarrow I_{y_g}$ maps a set of meaningful facial parameters to a rendered image. 
The facial parameter $y_{g}$ is a 269-dimensional vector with \fzl{the} head pose $h\in \mathbb R^{1\times2}$, AU parameters $y_{au} \in \mathbb R^{1\times23}$, and identity parameters $y_{id} \in \mathbb R^{1\times244}$. 
The identity parameter $y_{id}$ contains the continuous 208-dimensional parameters and a discrete parameter, which represents 36 styles of the brown.
The \sxh{rendered image is generated with the bone-driven face 3D model. In other words, it is the} ground truth $I_{y_g} \in \mathbb R^{H\times W\times 3}$.
\sxh{While, the generated image $I_{y_g}'$ is generated with a deep model (generator).}
Our generator $\mathcal{G}(y_g)$ consists of eight transposed convolution layers, which is similar to the generator of DCGAN \cite{radford2015unsupervised}.
We use instance normalization \cite{ulyanov2016instance} instead of batch normalization \cite{ioffe2015batch} for improving training stability in the generator, which is also used in GENet \cite{genet2020}.
We aim to minimize the difference between the rendered image and the generated one in the raw pixel space.
We adopt a content loss $\mathcal(L_{app})$ and a perceptual loss $\mathcal(L_{per})$ between the rendered image $I_{y_g}$ and the predicted one $I_{y_g}'=\mathcal{G}(y_g)$.
The losses are defined as:

\begin{equation}
\begin{aligned}
\mathcal{L}_{app}(y_g) &= E_{y_g\sim u(y_g)}[\Arrowvert I_{y_g}-I_{y_g}'\Arrowvert_{1}],
\end{aligned}
\label{l1loss}
\end{equation}

\begin{equation}
\begin{aligned}
\mathcal{L}_{per}(y_g) &= E_{y_g\sim u(y_g)}[\Arrowvert \mathcal{F}(I_{y_g}')-\mathcal{F}(I_{y_g})\Arrowvert_{2}].
\end{aligned}
\label{eq_perceputalloos}
\end{equation}
The two losses are combined to learn the generator $\mathcal{G}$ as:
\begin{equation} 
\mathcal{L}_{\mathcal{G}}(y_g) =\mathcal{L}_{app}(y_g) +\lambda\mathcal{L}_{per}(y_g),
\end{equation}
where $y_g$ is the facial parameters sampled from a uniform distribution $u(y_g)$. 
$\mathcal{F}$ denotes the \text{relu2\_2, relu3\_3, relu4\_3} feature maps in VGG16 \cite{simonyan2014very}, which is pre-trained on the image recognition task. 
\fzl{The $\lambda$ is the balance parameter}.

\begin{figure} 
	\includegraphics[width=0.96\linewidth]{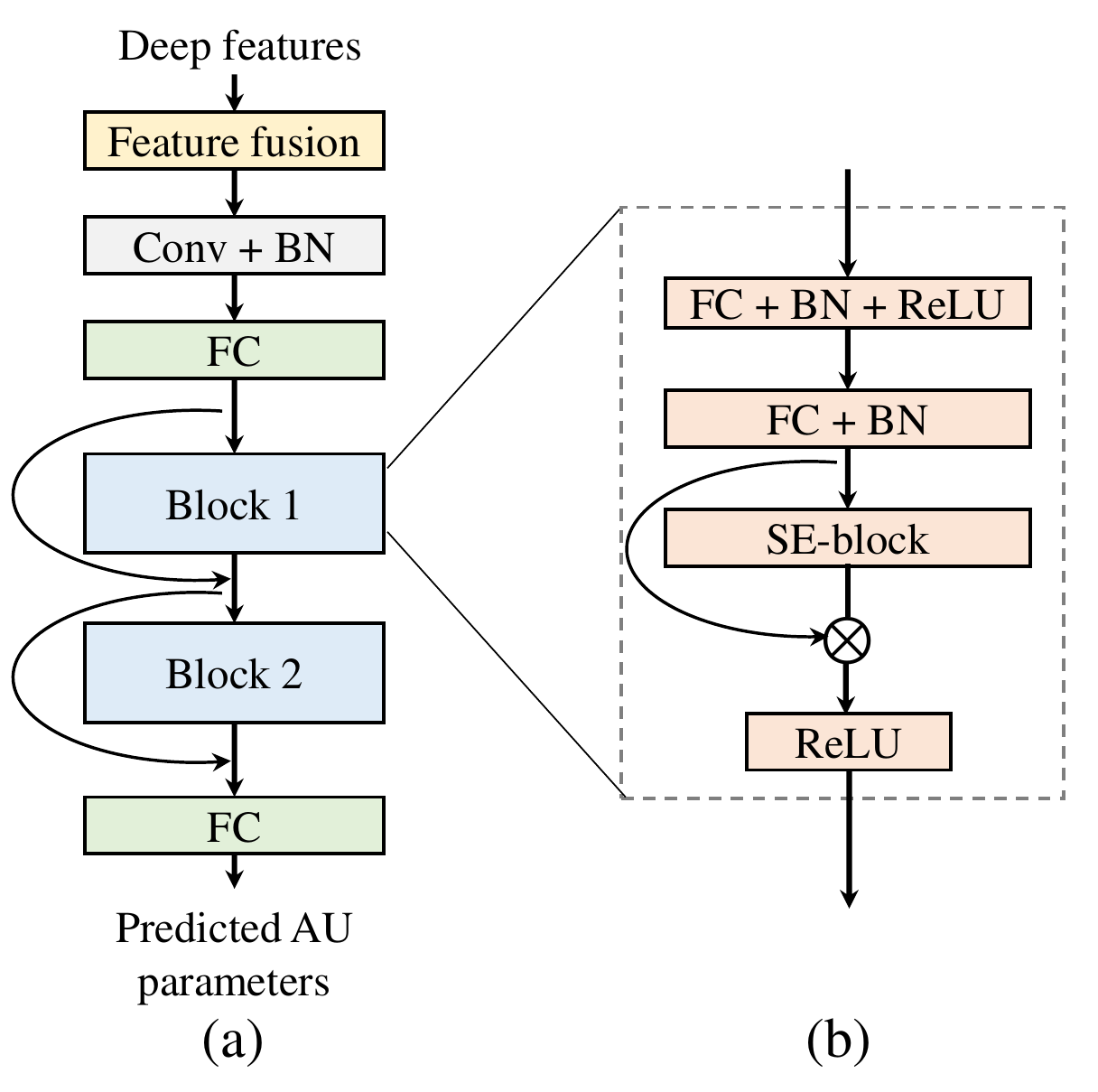}
	\caption{(a) The AU parameter regressive network. The regressive network consists of a feature fusion block, a convolution layer, two attention blocks and two fully connected (FC) layers. (b) The details of the attention block. $\bigotimes$ means element-wise product. SE-block is similar to the squeeze and excitation block in SENet \cite{hu2018squeeze} and it is applied to an FC layer. } 
	\Description{The AU parameters regressive network.}
	\label{regressnet}
\end{figure}

\subsection{Feature Extractor}
\label{featExt_sec}
\fzl{It's noticed that there is a large domain gap between the generated image and the real input image. To effectively measure the distance of the two images, we adopt a facial segmentation network as the feature extractor $F_{seg}$ for extracting the face features, which has been proved effective in GENet \cite{genet2020}.}
Specifically,
\fzl{The feature extractor $F_{seg}$ is based on the pretrained BiSeNet \cite{yu2018bisenet}\footnote{https://github.com/zllrunning/face-parsing.PyTorch} in our framework.}
$I$ is the input face image or the generated one, which is aligned via five facial landmarks using affine transformation, including left eye, right eye, nose, left mouth corner, and right mouth corner. The facial landmarks are obtained by OpenFace \cite{baltruvsaitis2015cross}.
The feature $f_{gb}$ \fzl{extracted} from the feature fusion module is used to describe the global information. And the latter three layers feature maps of ResNet18 \cite{he2016deep} are extracted to describe the local information in the context path of BiSeNet.
The feature extractor is defined as:
\begin{equation}
f_{gb}, T = F_{seg}(I),
\end{equation}
where $T$ is the set of the feature maps with weights.

\sxh{Furthermore,} an attention mechanism is deployed in the feature maps of the last three layers.
It allows the feature maps 
\fzl{to focus on} different regions of the face. 
The output probability maps of the facial segmentation network serve as the attention masks, which are represented as the pixel-wise weights $A$ in Eqn. (\ref{featExtract}).
For example, 
\fzl{the first layer's feature map} is sensitive to the browns, and the element-wise sum of the browns' probability maps is multiplied to the feature map of the first layer.
Suppose $C$ is the set of the feature maps. 
The facial representation $T$ is the set of multiple layers\fzl{'} output features, and one of them can be defined as:
\begin{equation}
\label{featExtract}
\mathcal{T}_i = \beta_{i} A_i(I) C_i(I),
\end{equation}
where $\beta$ represents the weight of each feature map $C_{i}$.
Specifically, the facial representation $T$ consists of five components, including eyes $\mathcal{T}_e$, nose $\mathcal{T}_n$, mouth $\mathcal{T}_m$, face $\mathcal{T}_f$ and brown $\mathcal{T}_{b}$.
The features of eyes and browns are \fzl{extracted} from the first layer.
The features of the second layer are sensitive to the nose and mouth.
The facial silhouette is 
\fzl{extracted} from the feature of the third layer.

\begin{figure*}[t]
	\includegraphics[width=0.90\linewidth]{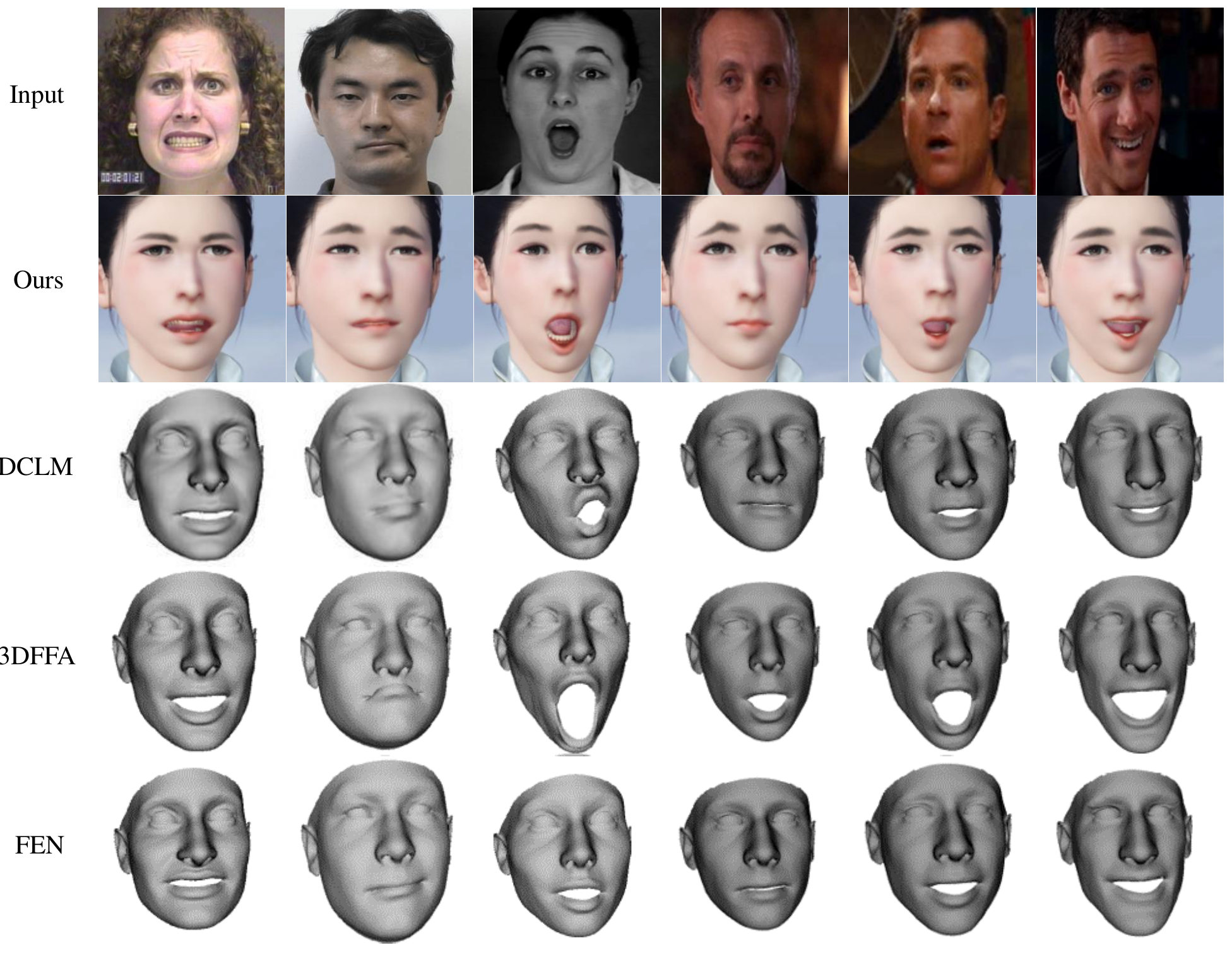}
	\caption{ Qualitative comparison with other methods. The results are shown in the same viewpoint. The results of DCLM \cite{zadeh2016deep}, 3DFFA \cite{zhu2016face} and FEN \cite{chang2019deep} are from original paper. The first three images are courtesy of FEN \cite{chang2019deep} in the first row. The last three original images are courtesy of Hector Elizondo, Jason Bateman, and Justin Bartha.
	} 
	\Description{Qualitative results.} 
	\label{qualitative_result}
\end{figure*}

\subsection{Facial Parameter Regressor}
The straightforward idea is that regressing all the facial parameters together through a fully connected layer.
However, each group of parameters has a strong meaning and responds to different features.  
The discovery has also been observed in multi-task learning work \cite{zhang2014facial}. 
Intuitively speaking, the head pose is not sensitive to local features since it is fundamentally independent of facial identity and subtle facial expressions. 
Whereas, for identity, both local and global features are necessary to distinguish the different persons.
The local and global features 
describe the shape of the four sense organs (including browns, eyes, nose, and mouth) and the overall facial silhouette, respectively.
Similarly, expression describing requires fine-grained features, such as mouth raised, pout and eye closed.
Therefore, we regress the facial parameters by different deep features, which are \fzl{extracted} from
the feature extractor (BiSeNet).
The feature $f_{gb}$ is used to regress the head pose. 
\fzl{The} facial representation $(\mathcal{T}_e,\mathcal{T}_n,\mathcal{T}_m,\mathcal{T}_f)$ is utilized to fit the continuous identity parameters and AU parameters. 
\fzl{The} $\mathcal{T}_{b}$ is used to regress the discrete identity parameters.
The predicted facial parameters $y'$ is defined as follows:

\begin{equation}
y' = \mathcal{R}(f_{gb}, T) = \mathcal{R}(F_{seg}(I)),
\end{equation}
where $I$ is \fzl{the} input face photo. 

\fzl{Specifically, the facial parameters regressor is separated into four sub-regressive networks: \fzl{the} head pose regressive network, the continuous identity parameters regressive network, the discrete identity parameters regressive network and the AU parameters regressive network. The four networks have similar network architecture, which is shown in Fig. \ref{regressnet}.}
A sub-regressive network contains a feature fusion block, a convolution layer, two attention blocks, and two fully connected (FC) layers. 
\fzl{We first append a global average pooling layer for different feature maps to resize the feature maps to the same size.}
And then concatenate the output features, called ``feature fusion". 
A convolution layer is performed with stride=2, followed by batch normalization.
The detail of the attention block is shown in Fig. \ref{regressnet} (b). Similar to the squeeze and excitation block in SENet \cite{hu2018squeeze}, we apply it to an FC layer, called ``SE-block".

\subsection{Losses}
We propose a novel loss function that jointly regresses the head pose, identity parameters, and AU parameters.
The loss function is conceptually straightforward and enables unsupervised training of our network. 
It combines five terms:
\begin{equation}
\mathcal{L} = \mathcal{L}_{ct} + w_{id}\mathcal{L}_{id} + w_{pr}\mathcal{L}_{pr} + w_{lp}\mathcal{}L_{lp}+ w_{adv}\mathcal{L}_{adv},
\end{equation}
where, $L_{ct}$ measures the pixel-wise distance of the facial representation from a pre-trained facial feature extractor,
$\mathcal{L}_{id}$ preserves the identity information on the real images,
$\mathcal{L}_{pr}$ is utilized to learn the sparse facial parameters,
$\mathcal{L}_{lp}$ ensures the network can correctly interpret the generated one,
and $\mathcal{L}_{adv}$ regularizes the predicted identity parameters.

\begin{figure*}[t]
	\includegraphics[width=0.98\linewidth]{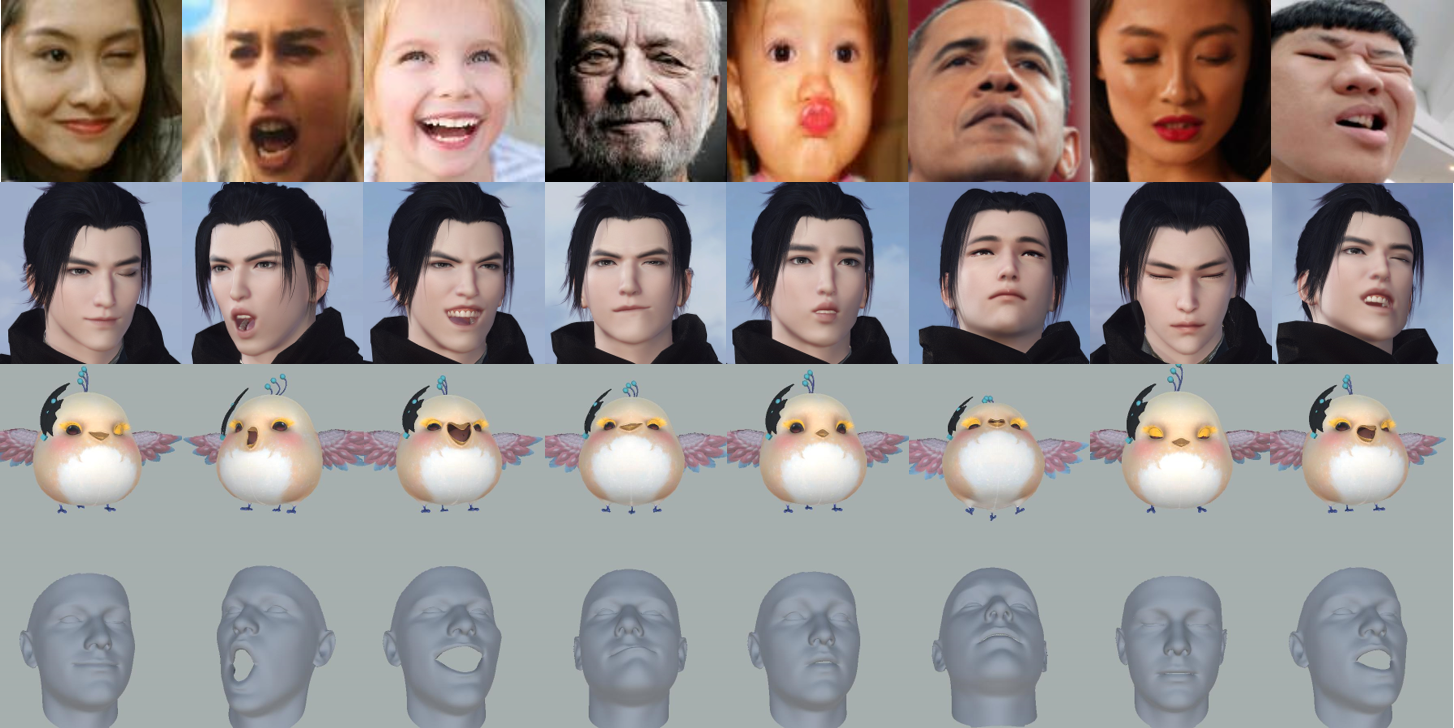}
	\caption{ Retargeting from faces to other 3D characters. The first row is the input images and the last three rows represent the results of transferring the head pose and AU parameters to other 3D characters. Original images courtesy of Yin Zhu, Emilia Clarke, pobhble 3ykn, Stephen Joshua Sondheim, Jackson, Barack Obam, HHFQ \cite{hhfq}, and Jackie Lin (ours), respectively.} 
	\Description{Retargeting from faces to other 3D characters.} 
	\label{transfer_result}
\end{figure*}
\textbf{Facial Content Loss.}
\fzl{To constrain the facial shape's similarity between the ``generated" result and the real input image, we define a facial content loss by computing the pixel-wise distance based on the facial representation $T$, which is introduced in Sec. \ref{featExt_sec}. }
The facial content loss $\mathcal{L}_{ct}$ is defined as:
\begin{equation}
\mathcal{L}_{ct} = \Arrowvert T_{g} - T_{r} \Arrowvert_{1} ,
\label{content_loss}
\end{equation} 
where $T_{g}$ and $T_{r}$ denote the facial representations of the generated and the real facial image, respectively.

\textbf{Identity Loss.}
To effectively disentangle the identity and AU parameters, we add the identity loss to measure the similarity between the input image and the emotionless generated one. 
The facial recognition network $F_{reg}$ named LightCNN-29v2 \cite{wu2018light}\footnote{https://github.com/AlfredXiangWu/LightCNN} \fzl{is adopted} to extract 256-d face embedding as the identity representation. 
The cosine similarity \cite{wang2018cosface} of two images is \fzl{set} as the identity loss, which is defined as:
\begin{equation}
\mathcal{L}_{id} = 1- \cos(m_{gbe}, m_r) = 1- \cos(F_{reg}(I_{y_{gbe}}), F_{reg}(I)),
\label{id_loss}
\end{equation}
where $m_{gbe}$ and $m_r$ are the identity embeddings of the emotionless generated image $I_{y_{gbe}}$ (is known as the generated image with ``base AU") and the real input image $I$ from the facial recognition network, respectively.
\fzl{The} ``base AU" represents all the AU parameters setting to 0.02.

\textbf{Parameter Loss.}
To the best of our knowledge, 
\fzl{there exists} the mutual exclusion of AUs in various facial expressions.
\fzl{For example, jaw left and jaw right, smile and lip stretcher can not occur at the same time.}
The same rule is applied to identity parameters. For an input image, we hope the predicted parameters are as sparsely as possible.
\fzl{The reason is that the dense parameters and the sparse parameters could have the same facial expression. For example, jaw left $0.5$ can be generated by (turning right: $0.4$, turning left: $0.9$). Meanwhile, it can also be generated by (turning right: $0$, turning left: $0.5$). So, some parameters can counteract the inverse facial parameters.
The parameter loss $\mathcal{L}_{pr}$ is adopted to enforce a sparsity constraint, which is defined as follows:}
\begin{equation}
\mathcal{L}_{pr} = \Arrowvert h' - h_{b}\Arrowvert^2_2 + \alpha \Arrowvert x_{id}' - x_{bid}\Arrowvert^2_2 + \beta \Arrowvert x_{au}' - x_{bau}\Arrowvert^2_2,
\label{param_loss}
\end{equation} 
\fzl{where, $h'$, $x_{id}'$ and $x_{au}'$ denote the predicted head pose, identity and AU parameters, respectively}.
\fzl{$h_{b}$, $x_{bid}$ and $x_{bau}$ denote} the frontal head pose, base identity parameters and emotionless AU parameters, which are set to $0.5$, $0.5$ and $0.02$, respectively.
\fzl{$\alpha$ and $\beta$ are the balance parameters.}

\begin{table*}[ht]
	\caption{The Intra-Class Correlation (ICC) and Mean Absolute Error (MAE) on BP4D and DISFA. Bold numbers indicate the best performance. Underline numbers indicate the second best. (*) indicates results taken from the reference. $\uparrow$ means the higher, the better. $\downarrow$ means the lower, the better.} 
	\label{compare_ICC_MAE}
	\begin{center} 
		\begin{tabular}{c|c|cccccc|ccccccccccccc} 
			\hline
			~& Database & \multicolumn{6}{c|}{BP4D}& \multicolumn{13}{c}{DISFA} \\
			\cline{2-21}
			~ & AU & 06 & 10 & 12& 14& 17 & Avg & 01 & 02& 04 &05& 06& 09& 12& 15 & 17& 20& 25 &26 & Avg \\
			\hline
			\multirow{6}{*}{{ICC}} & KBSS \cite{zhang2018weakly}* & .76 & {\bf.73} & .84 & .45 & .45 & .65         & .23 & .11 & .48 & .25 & .50 & .25 & .71 & .22 &.25 & .06& \underline{.83} &  .41 & .36 \\
			\multirow{7}{*}{{ $\uparrow$}} & KJRE \cite{zhang2019joint}* &.71 &.61 & {\bf.87} &.39 & .42 & .60 & .27 & .35 & .25 & .33 & .51 & .31 & .67 & .14 & .17 & .20 & .74 & .25 & .35  \\
			~ & LBA \cite{haeusser2017learning}* & .71 & .64 & .81 & .23 & .50  & .58  & .04 & .06 & .39 & .01 & .41 & .12 & \underline{.73} & .13 & .27& .10 & .82 & .43 & .29 \\
			~ & 2DC \cite{linh2017deepcoder}* &\underline{.76} & \underline{.71} & \underline{.85} & .45 & \underline{.53}  & \underline{.66} & {\bf.70} & .55 & \underline{.69} & .05 & {.59} & {.57} & {\bf.88} & .32 & .10 & .08 & {\bf.90} & .50 &  .50\\
			~ & CFLF \cite{zhang2019context}*& {\bf.77} & .70 & .83 & .41 & {\bf.60}  & \underline{.66} & .26 & .19 & .46 & .35 & .52 & .36 & .71 & .18 & .34 & .21 & .81 &.51 & .41\\ 
			~ &  GENet \cite{genet2020}*  & {.69} & {\bf.85} & .73 & {\bf.63} & \underline{.53}  & {\bf.68} & \underline{.66} & \underline{.67} & {\bf.73} & {\bf.71} & \underline{.60} & {\bf.59} & .60 & {\bf.66} & {\bf.58} & \underline{.45} &.80 &{\bf.70} & {\bf.64} \\
			~ &  Ours  & .67 & .62 & .67 & \underline{.62} & \underline{.53}  & .62 & 
			{.64} & {\bf.74} & {.59} & \underline{.67} & {\bf.63} & \underline{.58} & .55 & \underline{.61} & \underline{.49} & {\bf.50} &.71 &\underline{.66} & \underline{.61} \\
			\hline

			\multirow{6}{*}{{MAE }} & KBSS \cite{zhang2018weakly} *& .74 & \underline{.77} & .69 & .99 & .90  & .82  & .48 & .49 &  .57 & {\bf.08} & {\bf.26} & \underline{.22} & \underline{.33} &.15 & .44 & .22 & .43 & {.36} & .34  \\	
			\multirow{7}{*}{{ $\downarrow$}} & KJRE \cite{zhang2019joint}* & .82 & .95 & .64 & 1.08 & .85  &.87      & 1.0 & .92 & 1.9 & .70 & .79 & .87 & .77 & .60 & .80  & .72 & .96 & .94 & .91  \\	
			~ &LBA \cite{haeusser2017learning}* & .64 & .80 & .56 & 1.10 & {.62} & .74  &.43&  .29 &.51 &\underline{.10} &\underline{.30} &{\bf.19} &{\bf.30}& {\bf.11}& \underline{.31}& {\bf.14} &.40 & .38& {\bf.29} \\			
			
			~ & 2DC \cite{linh2017deepcoder}* & .87 & .84 & .92 & {.67} & .73  & .81 & .57 & .62 & .73 & .51 & .66 & .55 & .50 & .52 & .78 & .42 & .61 & .74 & .61\\
			
			~ & CFLF \cite{zhang2019context}* & \underline{.62} & .83 & .62 & 1.00 & .63  &.74 & {\bf.33} & {\bf.28} & .61 & .13 & .35 & .28 & .43 &.18 & {\bf.29} & \underline{.16} & .53 & .40 & \underline{.33} \\  
			~ &  GENet \cite{genet2020}* & .45 & .60 & {\bf.37} &{\bf.40} & {\bf.27}  & \underline{.42}  & \underline{.36} & {.33} & \textbf{.38} & .14 &.34 &.36 & .57 & .18 & \underline{.42} & .26& \textbf{.27} & \underline{.32} &\underline{.33}\\
			~ & Ours & {\bf.44} & {\bf.31} & \underline{.45} &\underline{.48} & \underline{.36}  & {\bf.41}  & {.40} & \underline{.21} & \underline{.39} & .16 &.28 &.30 & .51 &  \underline{.15} &.32 & .19 & \underline{.25} & {\bf.26} &{\bf.29}\\
			\hline
			
		\end{tabular}
	\end{center}
\end{table*}

\textbf{Loopback Loss.}
The true facial parameters of the real face images are unknown for unsupervised training. 
However, for the generated images, the true facial parameters are known. 
Inspired by the unsupervised 3D face reconstruction method proposed by Genova \etal~\cite{genova2018unsupervised}, 
\fzl{we introduce the ``loopback" loss, which constrains the consistency between the GT parameters and the predicted parameters of the image generated with the GT parameters}. 
\fzl{In the training stage, the extracted representations of the generated images $I_{gbe}$ and $I_{gbi}$ are fed into the facial parameter regressor to get the parameters $ y_{gbe}'$ and $y_{gbi}'$}.
$I_{gbe}$ and $I_{gbi}$ denote the generated images replaced by base AU and base identity.
$y_{gbe}$ and $y_{gbi}$ are the facial parameter ground truths with base AU or base identity. 
\fzl{So},  $y_{gbe} = (x_{bau}, x_{id}')$ and $y_{gbi} = (x_{au}', x_{bid})$.
The loopback loss is defined as follows:
\begin{equation}
\begin{aligned}
\mathcal{L}_{lp} &= \Arrowvert y_{gbe}- y_{gbe}' \Arrowvert_1 + \lambda_{le} \Arrowvert  y_{gbi} - y_{gbi}' \Arrowvert_1 \\
&=\Arrowvert y_{gbe}- \mathcal{R}(F_{seg}(I_{gbe}) \Arrowvert_1 + \lambda_{le} \Arrowvert  y_{gbi} - \mathcal{R}(F_{seg}(I_{gbi}) \Arrowvert_1, 
\label{loop_loss}
\end{aligned} 
\end{equation}
where $\lambda_{le}$ 
\fzl{is the balance parameter.}

\textbf{Adversarial Loss.}
\fzl{The bone-driven face model's identity parameters describe a large-scale 3D space where some points don't exist in the real world. }
To solve the problem, we adopt the adversarial loss to encourage the generated image indistinguishable from the ``base face" in the mouth region. 
It makes the AU parameters to response for the facial muscle movement. And the identity parameters keep the facial outline and the position of the four sense organs.
The ``base face" $I_{b}$ is created by rendering a frontal emotionless face image with the identity parameters setting to 0.5.
The facial representation of the ``base face" is represented to $\mathcal{T}_{bm}$ in the mouth region.
$I_{gbe}$ is the emotionless generated image. 
$\mathcal{T}_{gbem}$ represents the facial representation of the input images $I_{gbe}$ in the mouth region. 
\fzl{It is one group of the facial representation $T$ extracted from the feature extractor $F_{seg}$ in the mouth region.}
The adversarial loss is defined as:
\begin{equation}
\mathcal{L}_{adv} = \Arrowvert \mathcal{T}_{gbem} -\mathcal{T}_{bm}  \Arrowvert _1 
= \Arrowvert F_{seg}(I_{gbe}) - F_{seg}(I_{b})  \Arrowvert _1.
\label{adv_loss}
\end{equation}

\section{Experiments}

\subsection{Datasets}
We combine multiple datasets as a training set to 
\fzl{accurately} predict the facial parameters.
Facewarehouse \cite{cao2013facewarehouse} and MMI \cite{pantic2005web,valstar2010induced} are rich datasets for expressions.
DISFA \cite{mavadati2013disfa,mavadati2012automatic} and the Binghamton 4D (BP4D) \cite{zhang2014bp4d,zhang2016multimodal,valstar2017fera} are the two largest expression databases for AU intensity estimation.
Among them, DISFA contains 27 subjects, and the AU intensity of 12 AUs (1, 2, 4, 5, 6, 9, 12, 15, 17, 20, 25, 26) are annotated.
We use 18 subjects for training and 9 subjects for testing.
BP4D contains 41 subjects, and 5 AU intensity (6, 10, 12, 14, 17) are annotated.
We use 21 subjects for training and 20 subjects for testing.
The extended cohn-kanade dataset (CK+) \cite{lucey2010extended} contains 123 objects. 100 objects are used for training.
Due to the datasets are all frontal face images, so we augment them to multiple poses with the same identity and expression by the high-fidelity pose and expression normalization (HPEN) \cite{zhu2015high}. 
Besides, the expression distribution is 
\fzl{nonuniform} in the datasets.
So we augment the data through warping the images according to 68 key points.

\subsection{Training Set}
\fzl{We first train the facial generator with $80,000$ paired samples, which are constituted of the facial parameters and the corresponding rendered images.}
And then we fixed the network parameters of the proposed framework except \fzl{for} the facial parameter regressor. 
We set  $w_{id} = 0.1 , w_{pr} = 0.1 , w_{lp} = 0.1$ and $w_{adv}=0.1$ in our framework.
$\alpha$ and $\beta$ are set to $0.1$ and $0.01$.
$\lambda_{le}$ is set to $1$.

\subsection{Compared with Other Methods}

\textbf{Qualitative results.} 
We compare our method with several 3D face reconstruction methods, including DCLM \cite{zadeh2016deep}, 3DFFA \cite{zhu2016face} and FEN \cite{chang2019deep}. 
Fig. \ref{qualitative_result} provides rendered results with the same head pose for CK+ and EmotiW-17 \cite{dhall2017individual}.  
The second row shows the rendered results of the facial parameters estimated by our method.
The other three methods are based on 3DMM and their results are from the paper \cite{chang2019deep}.
These rendered images support that our qualitative results have better appearance than other methods.
\fzl{To evaluate our method's efficiency,} we also show more estimation results and retarget facial motion to other 3D characters in Fig. \ref{transfer_result}.
The input images are from the EmotionNet dataset and MS-SFN \cite{chaudhuri2019joint}.
The 3D characters have the same meaningful AU parameters with the bone-driven face model.
The results 
\fzl{demonstrate} that this approach can effectively disentangle 
facial parameters.
 
\begin{figure}
	\includegraphics[width=0.96\linewidth]{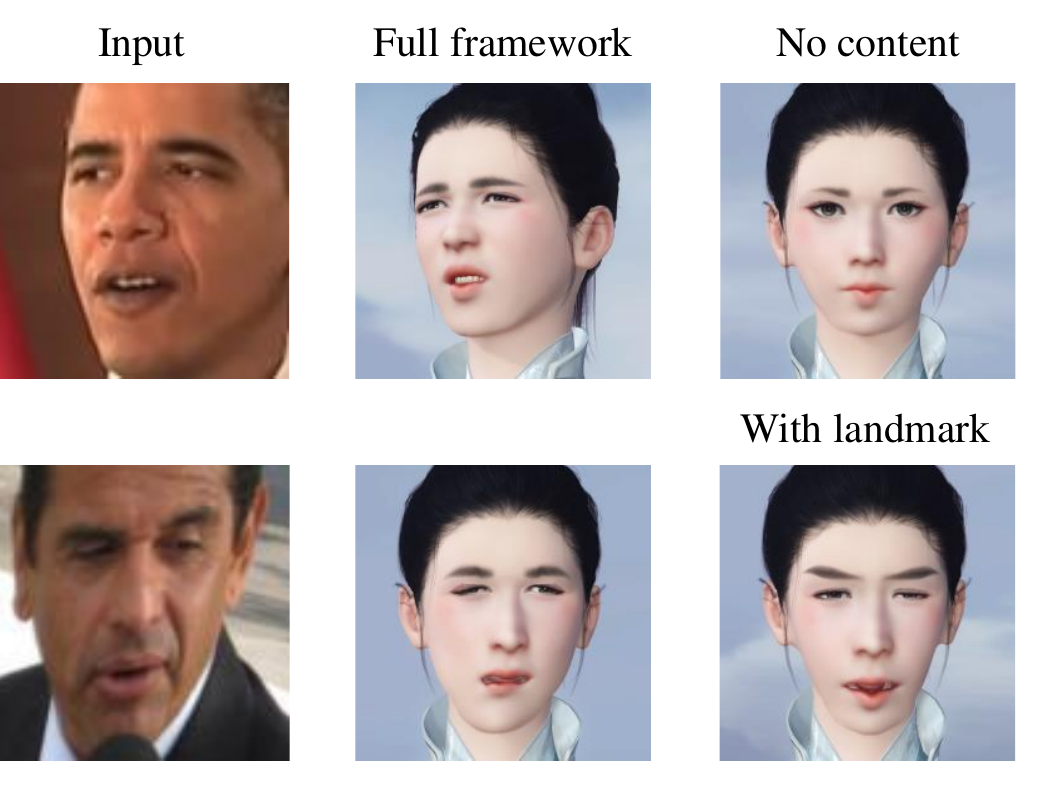}
	\caption{Ablation test showing the appearance by removing facial content loss (``no content") or using landmark loss (``with landmark") to replace. Column 1 corresponds to the input and column 2 corresponds to the results of our method, which is called ``full framework". The facial content loss is very important for keeping the results in the space of human faces. The facial landmark network can not capture the details of the facial appearance. Original images are courtesy of Barack Obama and Antonio Villaraigosa.
	} 
	\Description{Ablation test.} 
	\label{ablantion_qualitative}
\end{figure}
\textbf{Quantitative results.} 
\fzl{In this section, the DISFA and BP4D are adopted} to quantitatively evaluate \fzl{the} effectiveness of our method.
Intra-Class Correlation (ICC(3,1) \cite{shrout1979intraclass}) and Mean Absolute Error (MAE) are calculated to evaluate the AU intensity estimation methods between the ground truth and the predicted one.
We compare our method with several popular supervised learning methods, including ResNet18 \cite{he2016deep}, CCNN-IT \cite{walecki2017deep} and CNN \cite{gudi2015deep}.
CNN \cite{gudi2015deep} is composed of three convolutional layers and a fully connected layer. 
CCNN-IT \cite{walecki2017deep} takes advantage of CNNs and data augmentation. 
ResNet18 is introduced in \cite{he2016deep}. 
These supervised methods require annotated AU intensity of 
each frame in sequences while ours does not need the AU annotations.
Our method achieves better performance on both metrics (Table \ref{comparison_supervised}). 

We also compare the proposed method with several state-of-the-art weakly-supervised learning methods (KBSS \cite{zhang2018weakly}, KJRE \cite{zhang2019joint} and CFLF \cite{zhang2019context}), a semi-supervised learning method (LBA \cite{haeusser2017learning}) , a supervised method (2DC \cite{linh2017deepcoder}) and an unsupervised method (GENet \cite{genet2020}). 
LBA \cite{haeusser2017learning} estimates unlabeled samples with the assumption that samples with similar labels have similar latent features.
CFLF \cite{zhang2019context} uses \fzl{the} context-aware feature and label to train \fzl{a} patch-based deep model.
2DC \cite{haeusser2017learning} uses a semi-parametric VAE \fzl{framework}, while KBSS \cite{zhang2018weakly} and KJRE \cite{zhang2019joint} exploit domain knowledge to train a deep model.
Note that KBSS, KJRE, CFLF, LBA and 2DC need labeled data to train their models.
GENet \cite{genet2020} updates the facial parameters by iterative optimization process.
It takes 2 seconds to predict one image. 
Unlike them, we propose a framework for estimating AU intensity without annotated AU data.
And this proposed method can improve two orders of magnitude than GENet.
The quantitative results are listed in Table \ref{compare_ICC_MAE}.

On the BP4D and the DISFA datasets, our method can achieve superior average performance over other methods 
\fzl{with} MAE. 
On DISFA, our method achieves the second-best 
\fzl{with} ICC.
Note that ICC and MAE should be jointly considered to evaluate one method. 
LBA tends to predict the intensity to \fzl{be $0$}, which is the majority of AU intensity. It can achieve good MAE performance, but its ICC is much worse than ours. 
During the training phase, KBSS and CFLF need multiple frames as input; hence it requires more than one image per face, which is not always available.
These results show the superior performance of the proposed method over existing weakly and semi-supervised learning methods.
Besides, we also notice that our method has a bad performance than GENet on ICC. 
But our method can improve two orders of magnitude speed.

\subsection{Ablation Studies}
We perform several ablation experiments to verify the effectiveness of four loss functions in our framework.   
The facial content loss Eqn. (\ref{content_loss}) is the 
\fzl{critical} loss function in our method.
If no facial content loss, the facial parameter regressor can only learn the identity parameters 
\fzl{, not head pose or AU parameters.}
Beyond that, the facial landmark network is a popular basic task for describing facial features.
We also use the facial landmark network instead of the feature extractor.
Specifically, The last output heatmap and the features of the first two layers of the facial landmark network\footnote{https://github.com/1adrianb/face-alignment} are used to describe the global and local information, respectively.
And the facial content loss is replaced by MSE loss of the heatmaps.  
\fzl{Experimental results demonstrate} that the facial landmark network can not capture the detail information.
\fzl{Due to the significant differences between the results,}
so we only show server examples in Fig. \ref{ablantion_qualitative}. 
Besides, with the help of the identity loss Eqn. (\ref{id_loss}), our method can \fzl{noticeably} improve the results of the facial parameter regressor.  
For parameter loss Eqn. (\ref{param_loss}), adversarial loss Eqn. (\ref{adv_loss}) and loopback loss Eqn. (\ref{loop_loss}), we utilize the same evaluation metric as the quantitative comparison experiment.
The quantitative ablation results are shown in \fzl{Table} \ref{ablation_quantitive}.
They can help our method to have a significant improvement.

\begin{table} [t]
	\caption{Comparison with several supervised methods on DISFA. (*) indicates results taken from the reference.} 
	\begin{center}
		
		\begin{tabular}
			{ccccc}
			\toprule
			Method & CNN \cite{gudi2015deep}* & ResNet18 \cite{he2016deep}* &  CNN-IT \cite{walecki2017deep}* & Ours  \\
			\midrule
			ICC  & .33 & .27 & .38 & {\bf.61} \\
			MAE & .42 & .48 & .66 & {\bf.29}\\
			\bottomrule
		\end{tabular}
	\end{center}
	\label{comparison_supervised}
\end{table}

\begin{table}[t]
	\caption{Ablations analysis on different loss functions of the proposed method.}
	\begin{tabular}{cccc|cccc}
		\toprule
		\multicolumn{4}{c|}{Loss function } & \multicolumn{2}{c}{BP4D}  & \multicolumn{2}{c}{DISFA}  \\
		$\mathcal{L}_{id}$	& $\mathcal{L}_{lp}$ & $\mathcal{L}_{pr}$ & $\mathcal{L}_{adv}$& ICC & MAE & ICC & MAE \\
		\midrule
		$\times$  &  $\checkmark$ & $\checkmark$  & $\checkmark$     &  .53 & .65  & .55 & .35 \\
		$\checkmark$ & $\times$      & $\checkmark$  & $\checkmark$  & .60 & .64 & .52 & .37\\
		$\checkmark$ & $\checkmark$  & $\times$      & $\checkmark$  & .60 & .62 & .55 & .33 \\
		$\checkmark$ & $\checkmark$  & $\checkmark$  & $\times$      & .58 & .63 & .55 & .34\\
		\bottomrule
	\end{tabular}
	\label{ablation_quantitive}
\end{table}

\section{Conclusion}

This paper proposes \fzl{an} 
\sxh{unsupervised} learning framework for AU intensity estimation from a single face photo.
Different from the previous weakly-supervised methods and iterative optimization method, 
our method can estimate AU intensity without any annotated data. 
And the facial parameters regressor can improve 100x computational speedup than the iterative optimization method.
Quantitative and qualitative results have proved that the proposed method can achieve comparable results than other methods.
The ablation analysis also explains the effectiveness of our method on the five components.
 
\begin{acks}
This work is supported by National Natural Science Foundation of China (61976186), Key Research and Development Program of Zhejiang Province (2018C01004), and the Major Scientifc Research Project of Zhejiang Lab (No. 2019KD0AC01).
\end{acks}

\bibliographystyle{ACM-Reference-Format}
\bibliography{sample-base}


\begin{thebibliography}{62}


\ifx \showCODEN    \undefined \def \showCODEN     #1{\unskip}     \fi
\ifx \showDOI      \undefined \def \showDOI       #1{#1}\fi
\ifx \showISBNx    \undefined \def \showISBNx     #1{\unskip}     \fi
\ifx \showISBNxiii \undefined \def \showISBNxiii  #1{\unskip}     \fi
\ifx \showISSN     \undefined \def \showISSN      #1{\unskip}     \fi
\ifx \showLCCN     \undefined \def \showLCCN      #1{\unskip}     \fi
\ifx \shownote     \undefined \def \shownote      #1{#1}          \fi
\ifx \showarticletitle \undefined \def \showarticletitle #1{#1}   \fi
\ifx \showURL      \undefined \def \showURL       {\relax}        \fi
\providecommand\bibfield[2]{#2}
\providecommand\bibinfo[2]{#2}
\providecommand\natexlab[1]{#1}
\providecommand\showeprint[2][]{arXiv:#2}

\bibitem[\protect\citeauthoryear{Baltru{\v{s}}aitis, Mahmoud, and
  Robinson}{Baltru{\v{s}}aitis et~al\mbox{.}}{2015}]%
        {baltruvsaitis2015cross}
\bibfield{author}{\bibinfo{person}{Tadas Baltru{\v{s}}aitis},
  \bibinfo{person}{Marwa Mahmoud}, {and} \bibinfo{person}{Peter Robinson}.}
  \bibinfo{year}{2015}\natexlab{}.
\newblock \showarticletitle{Cross-dataset learning and person-specific
  normalisation for automatic action unit detection}. In
  \bibinfo{booktitle}{\emph{Automatic Face and Gesture Recognition}},
  Vol.~\bibinfo{volume}{6}. IEEE, \bibinfo{pages}{1--6}.
\newblock


\bibitem[\protect\citeauthoryear{Baltru{\v{s}}aitis, Robinson, and
  Morency}{Baltru{\v{s}}aitis et~al\mbox{.}}{2014}]%
        {baltruvsaitis2014continuous}
\bibfield{author}{\bibinfo{person}{Tadas Baltru{\v{s}}aitis},
  \bibinfo{person}{Peter Robinson}, {and} \bibinfo{person}{Louis-Philippe
  Morency}.} \bibinfo{year}{2014}\natexlab{}.
\newblock \showarticletitle{Continuous conditional neural fields for structured
  regression}. In \bibinfo{booktitle}{\emph{European Conference on Computer
  Vision}}. Springer, \bibinfo{pages}{593--608}.
\newblock


\bibitem[\protect\citeauthoryear{Blanz and Vetter}{Blanz and Vetter}{1999}]%
        {blanz1999morphable}
\bibfield{author}{\bibinfo{person}{Volker Blanz} {and} \bibinfo{person}{Thomas
  Vetter}.} \bibinfo{year}{1999}\natexlab{}.
\newblock \showarticletitle{A morphable model for the synthesis of 3D faces}.
  In \bibinfo{booktitle}{\emph{Computer Graphics and Interactive Techniques}}.
  \bibinfo{pages}{187--194}.
\newblock


\bibitem[\protect\citeauthoryear{Cao, Hou, and Zhou}{Cao et~al\mbox{.}}{2014}]%
        {cao2014displaced}
\bibfield{author}{\bibinfo{person}{Chen Cao}, \bibinfo{person}{Qiming Hou},
  {and} \bibinfo{person}{Kun Zhou}.} \bibinfo{year}{2014}\natexlab{}.
\newblock \showarticletitle{Displaced dynamic expression regression for
  real-time facial tracking and animation}.
\newblock \bibinfo{journal}{\emph{Transactions on graphics}}
  \bibinfo{volume}{33}, \bibinfo{number}{4} (\bibinfo{year}{2014}),
  \bibinfo{pages}{1--10}.
\newblock


\bibitem[\protect\citeauthoryear{Cao, Weng, Zhou, Tong, and Zhou}{Cao
  et~al\mbox{.}}{2013}]%
        {cao2013facewarehouse}
\bibfield{author}{\bibinfo{person}{Chen Cao}, \bibinfo{person}{Yanlin Weng},
  \bibinfo{person}{Shun Zhou}, \bibinfo{person}{Yiying Tong}, {and}
  \bibinfo{person}{Kun Zhou}.} \bibinfo{year}{2013}\natexlab{}.
\newblock \showarticletitle{Facewarehouse: A 3d facial expression database for
  visual computing}.
\newblock \bibinfo{journal}{\emph{Transactions on Visualization and Computer
  Graphics}} \bibinfo{volume}{20}, \bibinfo{number}{3} (\bibinfo{year}{2013}),
  \bibinfo{pages}{413--425}.
\newblock


\bibitem[\protect\citeauthoryear{Chang, Tran, Hassner, Masi, Nevatia, and
  Medioni}{Chang et~al\mbox{.}}{2019}]%
        {chang2019deep}
\bibfield{author}{\bibinfo{person}{Feng-Ju Chang}, \bibinfo{person}{Anh~Tuan
  Tran}, \bibinfo{person}{Tal Hassner}, \bibinfo{person}{Iacopo Masi},
  \bibinfo{person}{Ram Nevatia}, {and} \bibinfo{person}{G{\'e}rard Medioni}.}
  \bibinfo{year}{2019}\natexlab{}.
\newblock \showarticletitle{Deep, landmark-free FAME: Face alignment, modeling,
  and expression estimation}.
\newblock \bibinfo{journal}{\emph{International Journal of Computer Vision}}
  \bibinfo{volume}{127}, \bibinfo{number}{6-7} (\bibinfo{year}{2019}),
  \bibinfo{pages}{930--956}.
\newblock


\bibitem[\protect\citeauthoryear{Chaudhuri, Vesdapunt, and Wang}{Chaudhuri
  et~al\mbox{.}}{2019}]%
        {chaudhuri2019joint}
\bibfield{author}{\bibinfo{person}{Bindita Chaudhuri},
  \bibinfo{person}{Noranart Vesdapunt}, {and} \bibinfo{person}{Baoyuan Wang}.}
  \bibinfo{year}{2019}\natexlab{}.
\newblock \showarticletitle{Joint face detection and facial motion retargeting
  for multiple faces}. In \bibinfo{booktitle}{\emph{Computer Vision and Pattern
  Recognition}}. \bibinfo{pages}{9719--9728}.
\newblock


\bibitem[\protect\citeauthoryear{Dhall, Goecke, Ghosh, Joshi, Hoey, and
  Gedeon}{Dhall et~al\mbox{.}}{2017}]%
        {dhall2017individual}
\bibfield{author}{\bibinfo{person}{Abhinav Dhall}, \bibinfo{person}{Roland
  Goecke}, \bibinfo{person}{Shreya Ghosh}, \bibinfo{person}{Jyoti Joshi},
  \bibinfo{person}{Jesse Hoey}, {and} \bibinfo{person}{Tom Gedeon}.}
  \bibinfo{year}{2017}\natexlab{}.
\newblock \showarticletitle{From individual to group-level emotion recognition:
  Emotiw 5.0}. In \bibinfo{booktitle}{\emph{ACM International Conference on
  Multimodal Interaction}}. \bibinfo{pages}{524--528}.
\newblock


\bibitem[\protect\citeauthoryear{Friesen and Ekman}{Friesen and Ekman}{1978}]%
        {friesen1978facial}
\bibfield{author}{\bibinfo{person}{E Friesen} {and} \bibinfo{person}{Paul
  Ekman}.} \bibinfo{year}{1978}\natexlab{}.
\newblock \showarticletitle{Facial action coding system: a technique for the
  measurement of facial movement}.
\newblock \bibinfo{journal}{\emph{Palo Alto}}  \bibinfo{volume}{3}
  (\bibinfo{year}{1978}).
\newblock


\bibitem[\protect\citeauthoryear{Garrido, Zollh{\"o}fer, Casas, Valgaerts,
  Varanasi, P{\'e}rez, and Theobalt}{Garrido et~al\mbox{.}}{2016}]%
        {garrido2016reconstruction}
\bibfield{author}{\bibinfo{person}{Pablo Garrido}, \bibinfo{person}{Michael
  Zollh{\"o}fer}, \bibinfo{person}{Dan Casas}, \bibinfo{person}{Levi
  Valgaerts}, \bibinfo{person}{Kiran Varanasi}, \bibinfo{person}{Patrick
  P{\'e}rez}, {and} \bibinfo{person}{Christian Theobalt}.}
  \bibinfo{year}{2016}\natexlab{}.
\newblock \showarticletitle{Reconstruction of personalized 3D face rigs from
  monocular video}.
\newblock \bibinfo{journal}{\emph{Transactions on Graphics}}
  \bibinfo{volume}{35}, \bibinfo{number}{3} (\bibinfo{year}{2016}),
  \bibinfo{pages}{1--15}.
\newblock


\bibitem[\protect\citeauthoryear{Genova, Cole, Maschinot, Sarna, Vlasic, and
  Freeman}{Genova et~al\mbox{.}}{2018}]%
        {genova2018unsupervised}
\bibfield{author}{\bibinfo{person}{Kyle Genova}, \bibinfo{person}{Forrester
  Cole}, \bibinfo{person}{Aaron Maschinot}, \bibinfo{person}{Aaron Sarna},
  \bibinfo{person}{Daniel Vlasic}, {and} \bibinfo{person}{William~T Freeman}.}
  \bibinfo{year}{2018}\natexlab{}.
\newblock \showarticletitle{Unsupervised training for 3d morphable model
  regression}. In \bibinfo{booktitle}{\emph{Computer Vision and Pattern
  Recognition}}. \bibinfo{pages}{8377--8386}.
\newblock


\bibitem[\protect\citeauthoryear{Gudi, Tasli, Den~Uyl, and Maroulis}{Gudi
  et~al\mbox{.}}{2015}]%
        {gudi2015deep}
\bibfield{author}{\bibinfo{person}{Amogh Gudi}, \bibinfo{person}{H~Emrah
  Tasli}, \bibinfo{person}{Tim~M Den~Uyl}, {and} \bibinfo{person}{Andreas
  Maroulis}.} \bibinfo{year}{2015}\natexlab{}.
\newblock \showarticletitle{Deep learning based facs action unit occurrence and
  intensity estimation}. In \bibinfo{booktitle}{\emph{Automatic Face and
  Gesture Recognition}}, Vol.~\bibinfo{volume}{6}. IEEE, \bibinfo{pages}{1--5}.
\newblock


\bibitem[\protect\citeauthoryear{Haeusser, Mordvintsev, and Cremers}{Haeusser
  et~al\mbox{.}}{2017}]%
        {haeusser2017learning}
\bibfield{author}{\bibinfo{person}{Philip Haeusser}, \bibinfo{person}{Alexander
  Mordvintsev}, {and} \bibinfo{person}{Daniel Cremers}.}
  \bibinfo{year}{2017}\natexlab{}.
\newblock \showarticletitle{Learning by Association--A Versatile
  Semi-Supervised Training Method for Neural Networks}. In
  \bibinfo{booktitle}{\emph{Computer Vision and Pattern Recognition}}.
  \bibinfo{pages}{89--98}.
\newblock


\bibitem[\protect\citeauthoryear{He, Zhang, Ren, and Sun}{He
  et~al\mbox{.}}{2016}]%
        {he2016deep}
\bibfield{author}{\bibinfo{person}{Kaiming He}, \bibinfo{person}{Xiangyu
  Zhang}, \bibinfo{person}{Shaoqing Ren}, {and} \bibinfo{person}{Jian Sun}.}
  \bibinfo{year}{2016}\natexlab{}.
\newblock \showarticletitle{Deep residual learning for image recognition}. In
  \bibinfo{booktitle}{\emph{Computer Vision and Pattern Recognition}}.
  \bibinfo{pages}{770--778}.
\newblock


\bibitem[\protect\citeauthoryear{Hu, Shen, and Sun}{Hu et~al\mbox{.}}{2018}]%
        {hu2018squeeze}
\bibfield{author}{\bibinfo{person}{Jie Hu}, \bibinfo{person}{Li Shen}, {and}
  \bibinfo{person}{Gang Sun}.} \bibinfo{year}{2018}\natexlab{}.
\newblock \showarticletitle{Squeeze-and-excitation networks}. In
  \bibinfo{booktitle}{\emph{Computer Vision and Pattern Recognition}}.
  \bibinfo{pages}{7132--7141}.
\newblock


\bibitem[\protect\citeauthoryear{Ioffe and Szegedy}{Ioffe and Szegedy}{2015}]%
        {ioffe2015batch}
\bibfield{author}{\bibinfo{person}{Sergey Ioffe} {and}
  \bibinfo{person}{Christian Szegedy}.} \bibinfo{year}{2015}\natexlab{}.
\newblock \showarticletitle{Batch normalization: Accelerating deep network
  training by reducing internal covariate shift}.
\newblock \bibinfo{journal}{\emph{arXiv preprint arXiv:1502.03167}}
  (\bibinfo{year}{2015}).
\newblock


\bibitem[\protect\citeauthoryear{Jeni, Girard, Cohn, and De~La~Torre}{Jeni
  et~al\mbox{.}}{2013}]%
        {jeni2013continuous}
\bibfield{author}{\bibinfo{person}{L{\'a}szl{\'o}~A Jeni},
  \bibinfo{person}{Jeffrey~M Girard}, \bibinfo{person}{Jeffrey~F Cohn}, {and}
  \bibinfo{person}{Fernando De~La~Torre}.} \bibinfo{year}{2013}\natexlab{}.
\newblock \showarticletitle{Continuous au intensity estimation using localized,
  sparse facial feature space}. In \bibinfo{booktitle}{\emph{Automatic Face and
  Gesture Recognition}}. IEEE, \bibinfo{pages}{1--7}.
\newblock


\bibitem[\protect\citeauthoryear{Jiang, Wu, Chen, and Zhang}{Jiang
  et~al\mbox{.}}{2019}]%
        {jiang2019disentangled}
\bibfield{author}{\bibinfo{person}{Zi-Hang Jiang}, \bibinfo{person}{Qianyi Wu},
  \bibinfo{person}{Keyu Chen}, {and} \bibinfo{person}{Juyong Zhang}.}
  \bibinfo{year}{2019}\natexlab{}.
\newblock \showarticletitle{Disentangled representation learning for 3D face
  shape}. In \bibinfo{booktitle}{\emph{Computer Vision and Pattern
  Recognition}}. \bibinfo{pages}{11957--11966}.
\newblock


\bibitem[\protect\citeauthoryear{Kaltwang, Rudovic, and Pantic}{Kaltwang
  et~al\mbox{.}}{2012}]%
        {kaltwang2012continuous}
\bibfield{author}{\bibinfo{person}{Sebastian Kaltwang}, \bibinfo{person}{Ognjen
  Rudovic}, {and} \bibinfo{person}{Maja Pantic}.}
  \bibinfo{year}{2012}\natexlab{}.
\newblock \showarticletitle{Continuous pain intensity estimation from facial
  expressions}. In \bibinfo{booktitle}{\emph{International Symposium on Visual
  Computing}}. Springer, \bibinfo{pages}{368--377}.
\newblock


\bibitem[\protect\citeauthoryear{Kaltwang, Todorovic, and Pantic}{Kaltwang
  et~al\mbox{.}}{2015a}]%
        {kaltwang2015doubly}
\bibfield{author}{\bibinfo{person}{Sebastian Kaltwang}, \bibinfo{person}{Sinisa
  Todorovic}, {and} \bibinfo{person}{Maja Pantic}.}
  \bibinfo{year}{2015}\natexlab{a}.
\newblock \showarticletitle{Doubly sparse relevance vector machine for
  continuous facial behavior estimation}.
\newblock \bibinfo{journal}{\emph{Transactions on Pattern Analysis and Machine
  Intelligence}} (\bibinfo{year}{2015}), \bibinfo{pages}{1748--1761}.
\newblock


\bibitem[\protect\citeauthoryear{Kaltwang, Todorovic, and Pantic}{Kaltwang
  et~al\mbox{.}}{2015b}]%
        {kaltwang2015latent}
\bibfield{author}{\bibinfo{person}{Sebastian Kaltwang}, \bibinfo{person}{Sinisa
  Todorovic}, {and} \bibinfo{person}{Maja Pantic}.}
  \bibinfo{year}{2015}\natexlab{b}.
\newblock \showarticletitle{Latent trees for estimating intensity of facial
  action units}. In \bibinfo{booktitle}{\emph{Computer Vision and Pattern
  Recognition}}. \bibinfo{pages}{296--304}.
\newblock


\bibitem[\protect\citeauthoryear{Karras, Laine, and Aila}{Karras
  et~al\mbox{.}}{2019}]%
        {hhfq}
\bibfield{author}{\bibinfo{person}{Tero Karras}, \bibinfo{person}{Samuli
  Laine}, {and} \bibinfo{person}{Timo Aila}.} \bibinfo{year}{2019}\natexlab{}.
\newblock \showarticletitle{A Style-Based Generator Architecture for Generative
  Adversarial Networks}.
\newblock \bibinfo{journal}{\emph{Computer Vision and Pattern Recognition}}
  (\bibinfo{year}{2019}), \bibinfo{pages}{4401--4410}.
\newblock


\bibitem[\protect\citeauthoryear{Linh~Tran, Walecki, Eleftheriadis, Schuller,
  Pantic, et~al\mbox{.}}{Linh~Tran et~al\mbox{.}}{2017}]%
        {linh2017deepcoder}
\bibfield{author}{\bibinfo{person}{Dieu Linh~Tran}, \bibinfo{person}{Robert
  Walecki}, \bibinfo{person}{Stefanos Eleftheriadis}, \bibinfo{person}{Bjorn
  Schuller}, \bibinfo{person}{Maja Pantic}, {et~al\mbox{.}}}
  \bibinfo{year}{2017}\natexlab{}.
\newblock \showarticletitle{Deepcoder: Semi-parametric variational autoencoders
  for automatic facial action coding}. In
  \bibinfo{booktitle}{\emph{International Conference on Computer Vision}}.
  \bibinfo{pages}{3190--3199}.
\newblock


\bibitem[\protect\citeauthoryear{Lucey, Cohn, Kanade, Saragih, Ambadar, and
  Matthews}{Lucey et~al\mbox{.}}{2010}]%
        {lucey2010extended}
\bibfield{author}{\bibinfo{person}{Patrick Lucey}, \bibinfo{person}{Jeffrey~F
  Cohn}, \bibinfo{person}{Takeo Kanade}, \bibinfo{person}{Jason Saragih},
  \bibinfo{person}{Zara Ambadar}, {and} \bibinfo{person}{Iain Matthews}.}
  \bibinfo{year}{2010}\natexlab{}.
\newblock \showarticletitle{The extended cohn-kanade dataset (ck+): A complete
  dataset for action unit and emotion-specified expression}. In
  \bibinfo{booktitle}{\emph{Computer Vision and Pattern
  Recognition-Workshops}}. IEEE, \bibinfo{pages}{94--101}.
\newblock


\bibitem[\protect\citeauthoryear{Mavadati and Mahoor}{Mavadati and
  Mahoor}{2014}]%
        {mavadati2014temporal}
\bibfield{author}{\bibinfo{person}{S~Mohammad Mavadati} {and}
  \bibinfo{person}{Mohammad~H Mahoor}.} \bibinfo{year}{2014}\natexlab{}.
\newblock \showarticletitle{Temporal facial expression modeling for automated
  action unit intensity measurement}. In
  \bibinfo{booktitle}{\emph{International Conference on Pattern Recognition}}.
  IEEE, \bibinfo{pages}{4648--4653}.
\newblock


\bibitem[\protect\citeauthoryear{Mavadati, Mahoor, Bartlett, and
  Trinh}{Mavadati et~al\mbox{.}}{2012}]%
        {mavadati2012automatic}
\bibfield{author}{\bibinfo{person}{S~Mohammad Mavadati},
  \bibinfo{person}{Mohammad~H Mahoor}, \bibinfo{person}{Kevin Bartlett}, {and}
  \bibinfo{person}{Philip Trinh}.} \bibinfo{year}{2012}\natexlab{}.
\newblock \showarticletitle{Automatic detection of non-posed facial action
  units}. In \bibinfo{booktitle}{\emph{2012 19th IEEE International Conference
  on Image Processing}}. IEEE, \bibinfo{pages}{1817--1820}.
\newblock


\bibitem[\protect\citeauthoryear{Mavadati, Mahoor, Bartlett, Trinh, and
  Cohn}{Mavadati et~al\mbox{.}}{2013}]%
        {mavadati2013disfa}
\bibfield{author}{\bibinfo{person}{S~Mohammad Mavadati},
  \bibinfo{person}{Mohammad~H Mahoor}, \bibinfo{person}{Kevin Bartlett},
  \bibinfo{person}{Philip Trinh}, {and} \bibinfo{person}{Jeffrey~F Cohn}.}
  \bibinfo{year}{2013}\natexlab{}.
\newblock \showarticletitle{Disfa: A spontaneous facial action intensity
  database}.
\newblock \bibinfo{journal}{\emph{Transactions on Affective Computing}}
  \bibinfo{volume}{4}, \bibinfo{number}{2} (\bibinfo{year}{2013}),
  \bibinfo{pages}{151--160}.
\newblock


\bibitem[\protect\citeauthoryear{Ming, Bugeau, Rouas, and Shochi}{Ming
  et~al\mbox{.}}{2015}]%
        {ming2015facial}
\bibfield{author}{\bibinfo{person}{Zuheng Ming}, \bibinfo{person}{Aur{\'e}lie
  Bugeau}, \bibinfo{person}{Jean-Luc Rouas}, {and} \bibinfo{person}{Takaaki
  Shochi}.} \bibinfo{year}{2015}\natexlab{}.
\newblock \showarticletitle{Facial action units intensity estimation by the
  fusion of features with multi-kernel support vector machine}. In
  \bibinfo{booktitle}{\emph{Automatic Face and Gesture Recognition}},
  Vol.~\bibinfo{volume}{6}. IEEE, \bibinfo{pages}{1--6}.
\newblock


\bibitem[\protect\citeauthoryear{Pantic, Valstar, Rademaker, and Maat}{Pantic
  et~al\mbox{.}}{2005}]%
        {pantic2005web}
\bibfield{author}{\bibinfo{person}{Maja Pantic}, \bibinfo{person}{Michel
  Valstar}, \bibinfo{person}{Ron Rademaker}, {and} \bibinfo{person}{Ludo
  Maat}.} \bibinfo{year}{2005}\natexlab{}.
\newblock \showarticletitle{Web-based database for facial expression analysis}.
  In \bibinfo{booktitle}{\emph{International Conference on Multimedia and
  Expo}}. IEEE, \bibinfo{pages}{5--pp}.
\newblock


\bibitem[\protect\citeauthoryear{Paysan, Knothe, Amberg, Romdhani, and
  Vetter}{Paysan et~al\mbox{.}}{2009}]%
        {bfm09}
\bibfield{author}{\bibinfo{person}{P. Paysan}, \bibinfo{person}{R. Knothe},
  \bibinfo{person}{B. Amberg}, \bibinfo{person}{S. Romdhani}, {and}
  \bibinfo{person}{T. Vetter}.} \bibinfo{year}{2009}\natexlab{}.
\newblock \showarticletitle{A 3D Face Model for Pose and Illumination Invariant
  Face Recognition}.
\newblock \bibinfo{journal}{\emph{International Conference on Advanced Video
  and Signal based Surveillance for Security, Safety and Monitoring in Smart
  Environments}}.
\newblock


\bibitem[\protect\citeauthoryear{Radford, Metz, and Chintala}{Radford
  et~al\mbox{.}}{2015}]%
        {radford2015unsupervised}
\bibfield{author}{\bibinfo{person}{Alec Radford}, \bibinfo{person}{Luke Metz},
  {and} \bibinfo{person}{Soumith Chintala}.} \bibinfo{year}{2015}\natexlab{}.
\newblock \showarticletitle{Unsupervised representation learning with deep
  convolutional generative adversarial networks}.
\newblock \bibinfo{journal}{\emph{arXiv preprint arXiv:1511.06434}}
  (\bibinfo{year}{2015}).
\newblock


\bibitem[\protect\citeauthoryear{Richardson, Sela, and Kimmel}{Richardson
  et~al\mbox{.}}{2016}]%
        {richardson20163d}
\bibfield{author}{\bibinfo{person}{Elad Richardson}, \bibinfo{person}{Matan
  Sela}, {and} \bibinfo{person}{Ron Kimmel}.} \bibinfo{year}{2016}\natexlab{}.
\newblock \showarticletitle{3D face reconstruction by learning from synthetic
  data}. In \bibinfo{booktitle}{\emph{International Conference on 3D Vision}}.
  IEEE, \bibinfo{pages}{460--469}.
\newblock


\bibitem[\protect\citeauthoryear{Rudovic, Pavlovic, and Pantic}{Rudovic
  et~al\mbox{.}}{2014}]%
        {rudovic2014context}
\bibfield{author}{\bibinfo{person}{Ognjen Rudovic}, \bibinfo{person}{Vladimir
  Pavlovic}, {and} \bibinfo{person}{Maja Pantic}.}
  \bibinfo{year}{2014}\natexlab{}.
\newblock \showarticletitle{Context-sensitive dynamic ordinal regression for
  intensity estimation of facial action units}.
\newblock \bibinfo{journal}{\emph{Transactions on Pattern Analysis and Machine
  Intelligence}} \bibinfo{volume}{37}, \bibinfo{number}{5}
  (\bibinfo{year}{2014}), \bibinfo{pages}{944--958}.
\newblock


\bibitem[\protect\citeauthoryear{Sandbach, Zafeiriou, and Pantic}{Sandbach
  et~al\mbox{.}}{2013}]%
        {sandbach2013markov}
\bibfield{author}{\bibinfo{person}{Georgia Sandbach}, \bibinfo{person}{Stefanos
  Zafeiriou}, {and} \bibinfo{person}{Maja Pantic}.}
  \bibinfo{year}{2013}\natexlab{}.
\newblock \showarticletitle{Markov random field structures for facial action
  unit intensity estimation}. In \bibinfo{booktitle}{\emph{International
  Conference on Computer Vision Workshops}}. \bibinfo{pages}{738--745}.
\newblock


\bibitem[\protect\citeauthoryear{Sanyal, Bolkart, Feng, and Black}{Sanyal
  et~al\mbox{.}}{2019}]%
        {sanyal2019learning}
\bibfield{author}{\bibinfo{person}{Soubhik Sanyal}, \bibinfo{person}{Timo
  Bolkart}, \bibinfo{person}{Haiwen Feng}, {and} \bibinfo{person}{Michael~J
  Black}.} \bibinfo{year}{2019}\natexlab{}.
\newblock \showarticletitle{Learning to regress 3D face shape and expression
  from an image without 3D supervision}. In \bibinfo{booktitle}{\emph{Computer
  Vision and Pattern Recognition}}. \bibinfo{pages}{7763--7772}.
\newblock


\bibitem[\protect\citeauthoryear{Shi, Yuan, Fan, Zou, Shi, and Liu}{Shi
  et~al\mbox{.}}{2019}]%
        {F2p}
\bibfield{author}{\bibinfo{person}{Tianyang Shi}, \bibinfo{person}{Yi Yuan},
  \bibinfo{person}{Changjie Fan}, \bibinfo{person}{Zhengxia Zou},
  \bibinfo{person}{Zhenwei Shi}, {and} \bibinfo{person}{Yong Liu}.}
  \bibinfo{year}{2019}\natexlab{}.
\newblock \showarticletitle{Face-to-Parameter Translation for Game Character
  Auto-Creation}. In \bibinfo{booktitle}{\emph{International Conference on
  Computer Vision}}. \bibinfo{pages}{161--170}.
\newblock


\bibitem[\protect\citeauthoryear{Shrout and Fleiss}{Shrout and Fleiss}{1979}]%
        {shrout1979intraclass}
\bibfield{author}{\bibinfo{person}{Patrick~E Shrout} {and}
  \bibinfo{person}{Joseph~L Fleiss}.} \bibinfo{year}{1979}\natexlab{}.
\newblock \showarticletitle{Intraclass correlations: uses in assessing rater
  reliability.}
\newblock \bibinfo{journal}{\emph{Psychological bulletin}}
  \bibinfo{volume}{86}, \bibinfo{number}{2} (\bibinfo{year}{1979}),
  \bibinfo{pages}{420}.
\newblock


\bibitem[\protect\citeauthoryear{Simonyan and Zisserman}{Simonyan and
  Zisserman}{2014}]%
        {simonyan2014very}
\bibfield{author}{\bibinfo{person}{Karen Simonyan} {and}
  \bibinfo{person}{Andrew Zisserman}.} \bibinfo{year}{2014}\natexlab{}.
\newblock \showarticletitle{Very deep convolutional networks for large-scale
  image recognition}.
\newblock \bibinfo{journal}{\emph{arXiv preprint arXiv:1409.1556}}
  (\bibinfo{year}{2014}).
\newblock


\bibitem[\protect\citeauthoryear{Song, Shi, Shao, Yuan, Feng, and Fan}{Song
  et~al\mbox{.}}{2020}]%
        {genet2020}
\bibfield{author}{\bibinfo{person}{Xinhui Song}, \bibinfo{person}{Tianyang
  Shi}, \bibinfo{person}{Tianjia Shao}, \bibinfo{person}{Yi Yuan},
  \bibinfo{person}{Zunlei Feng}, {and} \bibinfo{person}{Changjie Fan}.}
  \bibinfo{year}{2020}\natexlab{}.
\newblock \showarticletitle{Unsupervised facial action unit intensity
  estimation via differentiable optimization}.
\newblock \bibinfo{journal}{\emph{arXiv preprint arXiv:2004.05908}}
  (\bibinfo{year}{2020}).
\newblock


\bibitem[\protect\citeauthoryear{Tewari, Zollh{\"o}fer, Garrido, Bernard, Kim,
  P{\'e}rez, and Theobalt}{Tewari et~al\mbox{.}}{2018}]%
        {tewari2018self}
\bibfield{author}{\bibinfo{person}{Ayush Tewari}, \bibinfo{person}{Michael
  Zollh{\"o}fer}, \bibinfo{person}{Pablo Garrido}, \bibinfo{person}{Florian
  Bernard}, \bibinfo{person}{Hyeongwoo Kim}, \bibinfo{person}{Patrick
  P{\'e}rez}, {and} \bibinfo{person}{Christian Theobalt}.}
  \bibinfo{year}{2018}\natexlab{}.
\newblock \showarticletitle{Self-supervised multi-level face model learning for
  monocular reconstruction at over 250 hz}. In
  \bibinfo{booktitle}{\emph{Computer Vision and Pattern Recognition}}.
  \bibinfo{pages}{2549--2559}.
\newblock


\bibitem[\protect\citeauthoryear{Tewari, Zollhofer, Kim, Garrido, Bernard,
  Perez, and Theobalt}{Tewari et~al\mbox{.}}{2017}]%
        {tewari2017mofa}
\bibfield{author}{\bibinfo{person}{Ayush Tewari}, \bibinfo{person}{Michael
  Zollhofer}, \bibinfo{person}{Hyeongwoo Kim}, \bibinfo{person}{Pablo Garrido},
  \bibinfo{person}{Florian Bernard}, \bibinfo{person}{Patrick Perez}, {and}
  \bibinfo{person}{Christian Theobalt}.} \bibinfo{year}{2017}\natexlab{}.
\newblock \showarticletitle{Mofa: Model-based deep convolutional face
  autoencoder for unsupervised monocular reconstruction}. In
  \bibinfo{booktitle}{\emph{International Conference on Computer Vision
  Workshops}}. \bibinfo{pages}{1274--1283}.
\newblock


\bibitem[\protect\citeauthoryear{Tuan~Tran, Hassner, Masi, and
  Medioni}{Tuan~Tran et~al\mbox{.}}{2017}]%
        {tuan2017regressing}
\bibfield{author}{\bibinfo{person}{Anh Tuan~Tran}, \bibinfo{person}{Tal
  Hassner}, \bibinfo{person}{Iacopo Masi}, {and} \bibinfo{person}{G{\'e}rard
  Medioni}.} \bibinfo{year}{2017}\natexlab{}.
\newblock \showarticletitle{Regressing robust and discriminative 3D morphable
  models with a very deep neural network}. In
  \bibinfo{booktitle}{\emph{Computer Vision and Pattern Recognition}}.
  \bibinfo{pages}{5163--5172}.
\newblock


\bibitem[\protect\citeauthoryear{Ulyanov, Vedaldi, and Lempitsky}{Ulyanov
  et~al\mbox{.}}{2016}]%
        {ulyanov2016instance}
\bibfield{author}{\bibinfo{person}{Dmitry Ulyanov}, \bibinfo{person}{Andrea
  Vedaldi}, {and} \bibinfo{person}{Victor Lempitsky}.}
  \bibinfo{year}{2016}\natexlab{}.
\newblock \showarticletitle{Instance normalization: The missing ingredient for
  fast stylization}.
\newblock \bibinfo{journal}{\emph{arXiv preprint arXiv:1607.08022}}
  (\bibinfo{year}{2016}).
\newblock


\bibitem[\protect\citeauthoryear{Valstar and Pantic}{Valstar and
  Pantic}{2010}]%
        {valstar2010induced}
\bibfield{author}{\bibinfo{person}{Michel Valstar} {and} \bibinfo{person}{Maja
  Pantic}.} \bibinfo{year}{2010}\natexlab{}.
\newblock \showarticletitle{Induced disgust, happiness and surprise: an
  addition to the mmi facial expression database}. In
  \bibinfo{booktitle}{\emph{Proc. 3rd Intern. Workshop on EMOTION (satellite of
  LREC): Corpora for Research on Emotion and Affect}}. Paris, France.,
  \bibinfo{pages}{65}.
\newblock


\bibitem[\protect\citeauthoryear{Valstar, S{\'a}nchez-Lozano, Cohn, Jeni,
  Girard, Zhang, Yin, and Pantic}{Valstar et~al\mbox{.}}{2017}]%
        {valstar2017fera}
\bibfield{author}{\bibinfo{person}{Michel~F Valstar}, \bibinfo{person}{Enrique
  S{\'a}nchez-Lozano}, \bibinfo{person}{Jeffrey~F Cohn},
  \bibinfo{person}{L{\'a}szl{\'o}~A Jeni}, \bibinfo{person}{Jeffrey~M Girard},
  \bibinfo{person}{Zheng Zhang}, \bibinfo{person}{Lijun Yin}, {and}
  \bibinfo{person}{Maja Pantic}.} \bibinfo{year}{2017}\natexlab{}.
\newblock \showarticletitle{Fera 2017-addressing head pose in the third facial
  expression recognition and analysis challenge}. In
  \bibinfo{booktitle}{\emph{Automatic Face \& Gesture Recognition}}. IEEE,
  \bibinfo{pages}{839--847}.
\newblock


\bibitem[\protect\citeauthoryear{Walecki, Pavlovic, Schuller, Pantic,
  et~al\mbox{.}}{Walecki et~al\mbox{.}}{2017}]%
        {walecki2017deep}
\bibfield{author}{\bibinfo{person}{Robert Walecki}, \bibinfo{person}{Vladimir
  Pavlovic}, \bibinfo{person}{Bj{\"o}rn Schuller}, \bibinfo{person}{Maja
  Pantic}, {et~al\mbox{.}}} \bibinfo{year}{2017}\natexlab{}.
\newblock \showarticletitle{Deep structured learning for facial action unit
  intensity estimation}. In \bibinfo{booktitle}{\emph{Computer Vision and
  Pattern Recognition}}. \bibinfo{pages}{3405--3414}.
\newblock


\bibitem[\protect\citeauthoryear{Walecki, Rudovic, Pavlovic, and
  Pantic}{Walecki et~al\mbox{.}}{2016}]%
        {walecki2016copula}
\bibfield{author}{\bibinfo{person}{Robert Walecki}, \bibinfo{person}{Ognjen
  Rudovic}, \bibinfo{person}{Vladimir Pavlovic}, {and} \bibinfo{person}{Maja
  Pantic}.} \bibinfo{year}{2016}\natexlab{}.
\newblock \showarticletitle{Copula ordinal regression for joint estimation of
  facial action unit intensity}. In \bibinfo{booktitle}{\emph{Computer Vision
  and Pattern Recognition}}. \bibinfo{pages}{4902--4910}.
\newblock


\bibitem[\protect\citeauthoryear{Wang, Wang, Zhou, Ji, Gong, Zhou, Li, and
  Liu}{Wang et~al\mbox{.}}{2018}]%
        {wang2018cosface}
\bibfield{author}{\bibinfo{person}{Hao Wang}, \bibinfo{person}{Yitong Wang},
  \bibinfo{person}{Zheng Zhou}, \bibinfo{person}{Xing Ji},
  \bibinfo{person}{Dihong Gong}, \bibinfo{person}{Jingchao Zhou},
  \bibinfo{person}{Zhifeng Li}, {and} \bibinfo{person}{Wei Liu}.}
  \bibinfo{year}{2018}\natexlab{}.
\newblock \showarticletitle{Cosface: Large margin cosine loss for deep face
  recognition}. In \bibinfo{booktitle}{\emph{Computer Vision and Pattern
  Recognition}}. \bibinfo{pages}{5265--5274}.
\newblock


\bibitem[\protect\citeauthoryear{Wu, He, Sun, and Tan}{Wu
  et~al\mbox{.}}{2018}]%
        {wu2018light}
\bibfield{author}{\bibinfo{person}{Xiang Wu}, \bibinfo{person}{Ran He},
  \bibinfo{person}{Zhenan Sun}, {and} \bibinfo{person}{Tieniu Tan}.}
  \bibinfo{year}{2018}\natexlab{}.
\newblock \showarticletitle{A light CNN for deep face representation with noisy
  labels}.
\newblock \bibinfo{journal}{\emph{Transactions on Information Forensics and
  Security}} \bibinfo{volume}{13}, \bibinfo{number}{11} (\bibinfo{year}{2018}),
  \bibinfo{pages}{2884--2896}.
\newblock


\bibitem[\protect\citeauthoryear{Yi, Li, Cao, Shen, Li, Wang, and Tai}{Yi
  et~al\mbox{.}}{2019}]%
        {yi2019mmface}
\bibfield{author}{\bibinfo{person}{Hongwei Yi}, \bibinfo{person}{Chen Li},
  \bibinfo{person}{Qiong Cao}, \bibinfo{person}{Xiaoyong Shen},
  \bibinfo{person}{Sheng Li}, \bibinfo{person}{Guoping Wang}, {and}
  \bibinfo{person}{Yu-Wing Tai}.} \bibinfo{year}{2019}\natexlab{}.
\newblock \showarticletitle{Mmface: A multi-metric regression network for
  unconstrained face reconstruction}. In \bibinfo{booktitle}{\emph{Computer
  Vision and Pattern Recognition}}. \bibinfo{pages}{7663--7672}.
\newblock


\bibitem[\protect\citeauthoryear{Yu, Wang, Peng, Gao, Yu, and Sang}{Yu
  et~al\mbox{.}}{2018}]%
        {yu2018bisenet}
\bibfield{author}{\bibinfo{person}{Changqian Yu}, \bibinfo{person}{Jingbo
  Wang}, \bibinfo{person}{Chao Peng}, \bibinfo{person}{Changxin Gao},
  \bibinfo{person}{Gang Yu}, {and} \bibinfo{person}{Nong Sang}.}
  \bibinfo{year}{2018}\natexlab{}.
\newblock \showarticletitle{Bisenet: Bilateral segmentation network for
  real-time semantic segmentation}. In \bibinfo{booktitle}{\emph{European
  Conference on Computer Vision}}. \bibinfo{pages}{325--341}.
\newblock


\bibitem[\protect\citeauthoryear{Zadeh, Baltru{\v{s}}aitis, and Morency}{Zadeh
  et~al\mbox{.}}{2016}]%
        {zadeh2016deep}
\bibfield{author}{\bibinfo{person}{Amir Zadeh}, \bibinfo{person}{Tadas
  Baltru{\v{s}}aitis}, {and} \bibinfo{person}{Louis-Philippe Morency}.}
  \bibinfo{year}{2016}\natexlab{}.
\newblock \showarticletitle{Deep constrained local models for facial landmark
  detection}.
\newblock \bibinfo{journal}{\emph{arXiv preprint arXiv:1611.08657}}
  \bibinfo{volume}{3}, \bibinfo{number}{5} (\bibinfo{year}{2016}),
  \bibinfo{pages}{6}.
\newblock


\bibitem[\protect\citeauthoryear{Zhang, Yin, Cohn, Canavan, Reale, Horowitz,
  Liu, and Girard}{Zhang et~al\mbox{.}}{2014b}]%
        {zhang2014bp4d}
\bibfield{author}{\bibinfo{person}{Xing Zhang}, \bibinfo{person}{Lijun Yin},
  \bibinfo{person}{Jeffrey~F Cohn}, \bibinfo{person}{Shaun Canavan},
  \bibinfo{person}{Michael Reale}, \bibinfo{person}{Andy Horowitz},
  \bibinfo{person}{Peng Liu}, {and} \bibinfo{person}{Jeffrey~M Girard}.}
  \bibinfo{year}{2014}\natexlab{b}.
\newblock \showarticletitle{Bp4d-spontaneous: a high-resolution spontaneous 3d
  dynamic facial expression database}.
\newblock \bibinfo{journal}{\emph{Image and Vision Computing}}
  \bibinfo{volume}{32}, \bibinfo{number}{10} (\bibinfo{year}{2014}),
  \bibinfo{pages}{692--706}.
\newblock


\bibitem[\protect\citeauthoryear{Zhang, Dong, Hu, and Ji}{Zhang
  et~al\mbox{.}}{2018a}]%
        {zhang2018weakly}
\bibfield{author}{\bibinfo{person}{Yong Zhang}, \bibinfo{person}{Weiming Dong},
  \bibinfo{person}{Bao-Gang Hu}, {and} \bibinfo{person}{Qiang Ji}.}
  \bibinfo{year}{2018}\natexlab{a}.
\newblock \showarticletitle{Weakly-supervised deep convolutional neural network
  learning for facial action unit intensity estimation}. In
  \bibinfo{booktitle}{\emph{Computer Vision and Pattern Recognition}}.
  \bibinfo{pages}{2314--2323}.
\newblock


\bibitem[\protect\citeauthoryear{Zhang, Jiang, Wu, Fan, and Ji}{Zhang
  et~al\mbox{.}}{2019a}]%
        {zhang2019context}
\bibfield{author}{\bibinfo{person}{Yong Zhang}, \bibinfo{person}{Haiyong
  Jiang}, \bibinfo{person}{Baoyuan Wu}, \bibinfo{person}{Yanbo Fan}, {and}
  \bibinfo{person}{Qiang Ji}.} \bibinfo{year}{2019}\natexlab{a}.
\newblock \showarticletitle{Context-Aware Feature and Label Fusion for Facial
  Action Unit Intensity Estimation With Partially Labeled Data}. In
  \bibinfo{booktitle}{\emph{International Conference on Computer Vision}}.
  \bibinfo{pages}{733--742}.
\newblock


\bibitem[\protect\citeauthoryear{Zhang, Wu, Dong, Li, Liu, Hu, and Ji}{Zhang
  et~al\mbox{.}}{2019b}]%
        {zhang2019joint}
\bibfield{author}{\bibinfo{person}{Yong Zhang}, \bibinfo{person}{Baoyuan Wu},
  \bibinfo{person}{Weiming Dong}, \bibinfo{person}{Zhifeng Li},
  \bibinfo{person}{Wei Liu}, \bibinfo{person}{Bao-Gang Hu}, {and}
  \bibinfo{person}{Qiang Ji}.} \bibinfo{year}{2019}\natexlab{b}.
\newblock \showarticletitle{Joint representation and estimator learning for
  facial action unit intensity estimation}. In
  \bibinfo{booktitle}{\emph{Computer Vision and Pattern Recognition}}.
  \bibinfo{pages}{3457--3466}.
\newblock


\bibitem[\protect\citeauthoryear{Zhang, Zhao, Dong, Hu, and Ji}{Zhang
  et~al\mbox{.}}{2018b}]%
        {zhang2018bilateral}
\bibfield{author}{\bibinfo{person}{Yong Zhang}, \bibinfo{person}{Rui Zhao},
  \bibinfo{person}{Weiming Dong}, \bibinfo{person}{Bao-Gang Hu}, {and}
  \bibinfo{person}{Qiang Ji}.} \bibinfo{year}{2018}\natexlab{b}.
\newblock \showarticletitle{Bilateral ordinal relevance multi-instance
  regression for facial action unit intensity estimation}. In
  \bibinfo{booktitle}{\emph{Computer Vision and Pattern Recognition}}.
  \bibinfo{pages}{7034--7043}.
\newblock


\bibitem[\protect\citeauthoryear{Zhang, Girard, Wu, Zhang, Liu, Ciftci,
  Canavan, Reale, Horowitz, Yang, et~al\mbox{.}}{Zhang et~al\mbox{.}}{2016}]%
        {zhang2016multimodal}
\bibfield{author}{\bibinfo{person}{Zheng Zhang}, \bibinfo{person}{Jeff~M
  Girard}, \bibinfo{person}{Yue Wu}, \bibinfo{person}{Xing Zhang},
  \bibinfo{person}{Peng Liu}, \bibinfo{person}{Umur Ciftci},
  \bibinfo{person}{Shaun Canavan}, \bibinfo{person}{Michael Reale},
  \bibinfo{person}{Andy Horowitz}, \bibinfo{person}{Huiyuan Yang},
  {et~al\mbox{.}}} \bibinfo{year}{2016}\natexlab{}.
\newblock \showarticletitle{Multimodal spontaneous emotion corpus for human
  behavior analysis}. In \bibinfo{booktitle}{\emph{Proceedings of the IEEE
  Conference on Computer Vision and Pattern Recognition}}.
  \bibinfo{pages}{3438--3446}.
\newblock


\bibitem[\protect\citeauthoryear{Zhang, Luo, Loy, and Tang}{Zhang
  et~al\mbox{.}}{2014a}]%
        {zhang2014facial}
\bibfield{author}{\bibinfo{person}{Zhanpeng Zhang}, \bibinfo{person}{Ping Luo},
  \bibinfo{person}{Chen~Change Loy}, {and} \bibinfo{person}{Xiaoou Tang}.}
  \bibinfo{year}{2014}\natexlab{a}.
\newblock \showarticletitle{Facial landmark detection by deep multi-task
  learning}. In \bibinfo{booktitle}{\emph{European Conference on Computer
  Vision}}. Springer, \bibinfo{pages}{94--108}.
\newblock


\bibitem[\protect\citeauthoryear{Zhao, Gan, Wang, and Ji}{Zhao
  et~al\mbox{.}}{2016}]%
        {zhao2016facial}
\bibfield{author}{\bibinfo{person}{Rui Zhao}, \bibinfo{person}{Quan Gan},
  \bibinfo{person}{Shangfei Wang}, {and} \bibinfo{person}{Qiang Ji}.}
  \bibinfo{year}{2016}\natexlab{}.
\newblock \showarticletitle{Facial expression intensity estimation using
  ordinal information}. In \bibinfo{booktitle}{\emph{Computer Vision and
  Pattern Recognition}}. \bibinfo{pages}{3466--3474}.
\newblock


\bibitem[\protect\citeauthoryear{Zhu, Lei, Liu, Shi, and Li}{Zhu
  et~al\mbox{.}}{2016}]%
        {zhu2016face}
\bibfield{author}{\bibinfo{person}{Xiangyu Zhu}, \bibinfo{person}{Zhen Lei},
  \bibinfo{person}{Xiaoming Liu}, \bibinfo{person}{Hailin Shi}, {and}
  \bibinfo{person}{Stan~Z Li}.} \bibinfo{year}{2016}\natexlab{}.
\newblock \showarticletitle{Face alignment across large poses: A 3d solution}.
  In \bibinfo{booktitle}{\emph{Computer Vision and Pattern Recognition}}.
  \bibinfo{pages}{146--155}.
\newblock


\bibitem[\protect\citeauthoryear{Zhu, Lei, Yan, Yi, and Li}{Zhu
  et~al\mbox{.}}{2015}]%
        {zhu2015high}
\bibfield{author}{\bibinfo{person}{Xiangyu Zhu}, \bibinfo{person}{Zhen Lei},
  \bibinfo{person}{Junjie Yan}, \bibinfo{person}{Dong Yi}, {and}
  \bibinfo{person}{Stan~Z Li}.} \bibinfo{year}{2015}\natexlab{}.
\newblock \showarticletitle{High-fidelity pose and expression normalization for
  face recognition in the wild}. In \bibinfo{booktitle}{\emph{Computer Vision
  and Pattern Recognition}}. \bibinfo{pages}{787--796}.
\newblock


\end{thebibliography}


\end{document}